\documentclass[12pt]{article}

\usepackage[spanish,activeacute]{babel}
\usepackage[T1]{fontenc}

\usepackage{amsmath, amsfonts, amssymb, amsthm}
\usepackage{bm}
\usepackage{mathrsfs}
\usepackage{dsfont}
\usepackage{scalerel} 

\usepackage{graphicx}
\usepackage{adjustbox}
\usepackage{grffile}
\usepackage{float}
\usepackage{flafter}
\usepackage{subfigure}
\usepackage{pdfpages}
\usepackage{pdflscape}

\usepackage{multirow}
\usepackage{multicol}
\usepackage{rotating}
\usepackage{threeparttable}

\usepackage{comment}
\usepackage{caption}
\usepackage{enumitem}
\usepackage{color, xcolor}
\usepackage{soul}
\usepackage{url}
\usepackage{setspace}
\usepackage[toc]{appendix}
\usepackage[hidelinks]{hyperref}
\usepackage{booktabs}

\usepackage[numbers]{natbib}
\setcitestyle{authoryear,open={(},close={)}}

\hypersetup{
    colorlinks=true,
    linkcolor=black,
    filecolor=black,
    urlcolor=black,
    citecolor=black
}

\def\spacingset#1{\renewcommand{\baselinestretch}{#1}\small\normalsize}
\spacingset{1}

\newcommand{\be}{\begin{eqnarray*}}
\newcommand{\ee}{\end{eqnarray*}}
\newcommand{\bet}{\begin{eqnarray}}
\newcommand{\eet}{\end{eqnarray}}

\def\@roman#1{\romannumeral #1}

\makeatletter
\def\hlinewd#1{%
  \noalign{\ifnum0=`}\fi\hrule \@height #1 %
  \futurelet\reserved@a\@xhline}
\makeatother

\begin{document}

\title{Construyendo la verdad: minería de texto y redes lingüísticas en audiencias públicas del Caso 03 de la Jurisdicción Especial para la Paz (JEP)}

\author{
    Juan Sosa$^1$\footnote{Contacto: jcsosam@unal.edu.co.}\quad
    Alejandro Urrego-López$^1$ \\
    Cesar Prieto$^1$ \quad
    Emma J. Camargo-Díaz$^2$ \\  
}	 

\date{
    $^1$Universidad Nacional de Colombia, Colombia \\ \vspace{0.3cm}
    $^2$Universidad Externado de Colombia, Colombia
}

\maketitle

\begin{abstract}
El Caso 03 de la Jurisdicción Especial para la Paz (JEP), centrado en los denominados \textit{falsos positivos} en Colombia, constituye uno de los episodios más desgarradores del conflicto armado colombiano. Este artículo propone una metodología innovadora basada en el análisis de lenguaje natural y modelos de co-ocurrencia semántica para explorar, sistematizar y visualizar patrones narrativos presentes en las audiencias públicas de víctimas y comparecientes. A través de la construcción de redes de skipgramas y el estudio de su modularidad, se identifican clústeres temáticos que revelan diferencias regionales y de estatus procesal, aportando evidencia empírica sobre dinámicas de victimización, responsabilidad y reconocimiento en este caso. Este enfoque computacional permite avanzar en la construcción colectiva de la verdad judicial y extrajudicial, ofreciendo herramientas replicables para otros casos de justicia transicional. El trabajo se apalanca en los pilares de verdad, justicia, reparación y no repetición, proponiendo una lectura crítica y profunda de las memorias en disputa.
\end{abstract}

\noindent
{\it Palabras clave: Justicia transicional; Análisis de lenguaje natural; Redes semánticas; JEP; Narrativas del conflicto armado.}

\spacingset{1.1} 

\newpage

\section{Introducción}

La situación conocida como los \textit{falsos positivos} en Colombia tuvo lugar en el prolongado conflicto armado entre el Estado, grupos guerrilleros como las Fuerzas Armadas Revolucionarias de Colombia (FARC-EP) y el Ejército de Liberación Nacional (ELN), así como organizaciones paramilitares. A partir de la década del 2000, durante el gobierno de Álvaro Uribe Vélez (2002--2010), se implementaron las políticas de Seguridad Democrática, mediante las cuales las Fuerzas Armadas enfrentaron una fuerte presión para demostrar resultados en la lucha contra los grupos armados ilegales \citep{Ruiz2021}. El éxito militar se evaluaba, principalmente, por el número de bajas en combate, lo que incentivó prácticas como la ejecución de civiles presentados falsamente como guerrilleros muertos en enfrentamientos \citep{Dejusticia2022}.

Una manifestación concreta de esta lógica fue la Directiva 029, emitida por el Ministerio de Defensa en 2005, que promovió recompensas sin distinguir entre capturas y bajas. Esta política derivó, en numerosos casos, en la ejecución de civiles —en su mayoría jóvenes de sectores vulnerables— quienes fueron presentados como combatientes abatidos con el fin de obtener beneficios institucionales, ascensos o reconocimientos \citep{ComisionVerdad2022}. La ausencia de controles eficaces, la permisividad de altos mandos y la deshumanización del enemigo facilitaron la persistencia de esta práctica, convirtiendo los falsos positivos en uno de los episodios más oscuros del conflicto armado colombiano \citep{JEP2018}.

Diversas investigaciones judiciales y reportes de derechos humanos han documentado esta práctica de manera sistemática. Según fuentes judiciales y periodísticas, los hechos seguían un \textit{modus operandi} reiterado: las víctimas, generalmente jóvenes de zonas rurales o urbanas marginadas, eran reclutadas mediante engaños —frecuentemente con falsas promesas de empleo—, trasladadas a regiones apartadas, ejecutadas extrajudicialmente, y sus cuerpos posteriormente manipulados con armas y uniformes para simular bajas enemigas \citep{HRW2015, CPI2012, ElEspectador2021}.

La Jurisdicción Especial para la Paz (JEP) fue creada en el marco del Acuerdo de Paz firmado en 2016 entre el gobierno colombiano y las FARC-EP, con el objetivo de investigar, juzgar y sancionar los crímenes más graves cometidos durante el conflicto armado. Este tribunal hace parte del Sistema Integral de Verdad, Justicia, Reparación y No Repetición, diseñado para garantizar los derechos de las víctimas a través de mecanismos de verdad y justicia \citep{JEP2017}. Formalmente establecida en 2017, la JEP tiene el mandato de abordar crímenes de guerra, delitos de lesa humanidad y violaciones graves de derechos humanos cometidos antes del 1 de diciembre de 2016, priorizando sanciones restaurativas para quienes reconozcan responsabilidad y colaboren activamente con el esclarecimiento de los hechos \citep{ASCOA2018}.

El funcionamiento de la JEP ha estado acompañado de un intenso debate político y jurídico. Mientras algunos sectores cuestionan su legitimidad y eficacia \citep{Moreno2024}, otros la consideran un componente clave en la superación del conflicto armado \citep{AmbosAboueldahab2020}. La JEP ha logrado avances significativos en la identificación y priorización de casos emblemáticos como el de los falsos positivos, promoviendo la participación activa de las víctimas. Actualmente, se investigan 11 macrocasos, entre ellos el denominado ``Caso 03'', denominado "Asesinatos y desapariciones forzadas presentados como bajas en combate por agentes del Estado" práctica que se conoció en el país como "Falsos Positivos" \citep{JEP_Caso03}.

En el desarrollo del Caso 03, la JEP ha recopilado información de múltiples fuentes institucionales, como la Fiscalía General de la Nación, la Procuraduría, el Sistema Penal Acusatorio, y organizaciones de la sociedad civil como el Centro Nacional de Memoria Histórica y la Coordinación Colombia Europa Estados Unidos (CCEEU). Gracias a estos insumos, se han identificado más de 6.402 víctimas entre 2002 y 2008 \citep{JEP2023}. Entre las acciones principales se destacan la recolección de versiones voluntarias, el análisis de pruebas y la priorización de regiones con mayores niveles de victimización, como Antioquia, Norte de Santander, la Costa Caribe, Meta y Casanare. Hasta la fecha, más de 25 miembros de las Fuerzas Armadas y un civil han sido imputados por crímenes de guerra y de lesa humanidad, reconociendo su responsabilidad en audiencias públicas \citep{JEP2023b}.

Actualmente, el caso se encuentra en la etapa de imposición de sanciones propias para quienes han colaborado ampliamente con la verdad. La Sala de Reconocimiento ha emitido las primeras Resoluciones de Conclusiones, trasladando los expedientes al Tribunal para la Paz, que será el encargado de imponer las sanciones. Paralelamente, la Unidad de Investigación y Acusación (UIA) continúa indagando sobre altos mandos que no han reconocido responsabilidad, quienes podrían enfrentar penas de hasta 20 años de prisión \citep{JEP2022, Dejusticia2022}.

Con el propósito de contribuir a los pilares de Verdad, Justicia, Reparación y No Repetición, este estudio emplea técnicas de procesamiento del lenguaje natural (PLN) para analizar las declaraciones recabadas en las audiencias públicas del Caso 03. A través de un análisis sistemático de los relatos de víctimas y comparecientes, se busca reconstruir los hechos de manera objetiva e independiente, y proponer metodologías innovadoras para abordar conflictos caracterizados por múltiples narrativas.

La JEP ha puesto a disposición del público, a través de su canal de YouTube (\url{https://www.youtube.com/@jepcolombia}), los videos correspondientes a todas las audiencias públicas del Caso 03. En estas diligencias se recopilaron testimonios de miembros de las Fuerzas Armadas involucrados y de víctimas reconocidas. Este material constituye la fuente primaria de análisis de este trabajo. Para acceder a la totalidad de las audiencias, un ciudadano debería revisar aproximadamente 78 días, 18 horas, 43 minutos y 3 segundos de video, además de 580 páginas de documentos oficiales de la JEP. La estimación de la duración fue realizada con la herramienta Playlist Length (\url{https://www.playlistlength.com}), que consulta directamente la API de YouTube. Los documentos están disponibles en \url{https://www.jep.gov.co/Notificaciones/Forms/Todos%20los%20documentos_2.aspx}, donde pueden ser explorados por nombre o tema.

Este estudio presenta un análisis exhaustivo de dichas audiencias y de las conclusiones derivadas, con el objetivo de facilitar el acceso a la información, promover la comprensión pública del caso y contribuir al esclarecimiento de la verdad mediante métodos cuantitativos que fortalezcan los mecanismos de justicia transicional.

El análisis de las declaraciones se desarrolla desde cuatro perspectivas principales:
\begin{enumerate}
    \item Un estudio general del conjunto completo de datos recopilados.
    \item Un análisis desagregado por las regiones priorizadas por la JEP: Antioquia, Huila, Norte de Santander, Costa Caribe, Meta y Casanare.
    \item Una comparación entre las versiones de los sujetos imputados y las de las víctimas.
    \item Un análisis específico de los testimonios de miembros de las Fuerzas Armadas, enfocado en identificar el \textit{modus operandi} en la ejecución de los falsos positivos.
\end{enumerate}

El análisis de texto mediante redes se ha consolidado como una herramienta eficaz para modelar relaciones entre palabras, conceptos y actores \citep{luque2024caracterizacion}. En el contexto del Caso 03, esta metodología permite identificar patrones narrativos, relaciones temáticas y conexiones discursivas entre los distintos testimonios presentados ante la JEP. El uso de redes semánticas y de coocurrencia facilita la visualización de la frecuencia y centralidad de conceptos clave como ``ejecución extrajudicial'', ``recompensas'', ``bajas en combate'' y ``responsabilidad''.

El presente artículo se organiza de la siguiente manera: en la Sección 2 se realiza una revisión exhaustiva de la literatura relacionada con los falsos positivos y la justicia transicional en Colombia, destacando los principales hallazgos y enfoques previos. La Sección 3 describe la metodología empleada, incluyendo el marco teórico, las herramientas de análisis de texto y los criterios estadísticos utilizados. En la Sección 4 se presentan los resultados del análisis, divididos por subcasos regionales y diferenciados entre víctimas y comparecientes, incorporando visualizaciones, pruebas de hipótesis y redes semánticas. Finalmente, la Sección 5 ofrece las conclusiones principales del estudio, reflexionando sobre las implicaciones de los hallazgos y su contribución a los mecanismos de justicia transicional en Colombia.

\section{Revisión de literatura}

Esta sección presenta una revisión crítica de la literatura académica y documental sobre los falsos positivos en Colombia y el desarrollo de la justicia transicional. Se abordan las principales contribuciones desde el ámbito nacional e internacional, resaltando enfoques históricos, jurídicos y sociales, así como los avances recientes en materia de verdad, reparación y no repetición. Esta revisión permite contextualizar el fenómeno dentro del conflicto armado colombiano y fundamentar el enfoque analítico adoptado en este estudio.

\subsection{Falsos positivos}

La problemática de los falsos positivos en Colombia ha sido objeto de un amplio corpus académico que aborda el fenómeno desde distintas disciplinas, especialmente los estudios en derechos humanos, sociología, ciencia política y justicia transicional. Los primeros trabajos se enfocaron en la documentación exhaustiva de los hechos, estableciendo la magnitud de las violaciones cometidas por agentes del Estado, en particular por miembros de las Fuerzas Armadas. Estas investigaciones pioneras fueron impulsadas principalmente por organizaciones de derechos humanos y centros académicos como el Centro de Investigación y Educación Popular (CINEP) y el Centro Nacional de Memoria Histórica (CNMH).

Las publicaciones de estas instituciones no solo sistematizan casos individuales, sino que analizan también la responsabilidad estructural del Estado y las condiciones sociopolíticas que facilitaron la comisión de estas acciones. En la revista Noche y Niebla del CINEP se registran miles de casos de víctimas ocurridos entre 1984 y 2011, mientras que los informes del CNMH examinan cómo la política de incentivos dentro de las Fuerzas Armadas —como parte de la estrategia de Seguridad Democrática— propició la sistematización de estas prácticas, destacando las lógicas institucionales y los vacíos de control interno que las hicieron posibles.

En el plano internacional, diversos estudios han analizado el fenómeno de los falsos positivos desde el enfoque del derecho internacional humanitario y la justicia penal internacional. Investigaciones publicadas en revistas como Human Rights Quarterly y The International Journal of Transitional Justice han subrayado la gravedad de estas ejecuciones extrajudiciales como crímenes de guerra y de lesa humanidad. Entre los aportes destacados se encuentra el trabajo de \citet{Tate2015}, \textit{Drugs, Thugs, and Diplomats}, donde se examina el papel del Estado colombiano en la producción de violencia, así como el impacto de la política de seguridad en la sistematización de los falsos positivos. Asimismo, académicos como Nazih Richani han profundizado en los vínculos entre la militarización del conflicto y la violencia contra la población civil, denunciando la instrumentalización de las bajas como métrica de éxito operativo.

Más recientemente, la atención académica ha girado hacia las narrativas de las víctimas y su papel central en los procesos de verdad, memoria y justicia. Este enfoque ha enfatizado la necesidad de reconocer a las víctimas como sujetos políticos activos en la reconstrucción de la verdad histórica. Según \citet{Jakobsen2024}, la JEP ha sido clave en este proceso, al ofrecer un espacio institucional para que las víctimas puedan relatar sus experiencias y obtener reconocimiento público de sus sufrimientos. Las audiencias públicas han generado condiciones para el reconocimiento de responsabilidades por parte de los perpetradores, lo que ha fortalecido la legitimidad del proceso judicial y ha contribuido a una mayor confianza en los mecanismos de justicia transicional.

Además, diversos estudios han resaltado que la JEP no solo promueve la verdad y la justicia, sino que incorpora un enfoque integral de reparación, articulado con medidas de rehabilitación, satisfacción simbólica y garantías de no repetición. De acuerdo con \citet{BarbosaCiro2020}, uno de los principales desafíos que enfrenta esta jurisdicción es garantizar una justicia restaurativa que reconozca a las víctimas, repare sus derechos y transforme las condiciones estructurales que permitieron estos crímenes. En esta tarea, la JEP actúa de manera complementaria con la Comisión para el Esclarecimiento de la Verdad, desarrollando una base empírica sólida sobre las dinámicas institucionales que favorecieron la impunidad, como lo ha señalado también \citet{Tate2015}.

El fenómeno de los falsos positivos —entendido como la ejecución de civiles inocentes que luego eran presentados como insurgentes abatidos en combate— forma parte de un patrón más amplio de violencia estatal y violaciones sistemáticas a los derechos humanos. \citet{Acemoglu2020} argumentan que los incentivos desmedidos dentro de la estructura militar, al priorizar las bajas como indicador de éxito, generaron un entorno propicio para la normalización de estas prácticas, reflejando cómo las políticas de seguridad democrática contribuyeron a legitimar conductas abusivas.

La literatura revisada incluye aportes sustantivos sobre el impacto de los falsos positivos en los derechos humanos y la responsabilidad del Estado. \citet{Amnistia2002, Amnistia2008} advierten sobre los riesgos asociados a la asistencia militar de Estados Unidos y a las políticas de seguridad que favorecieron la impunidad. \citet{Angulo2012} reflexiona sobre la huella profunda que dejaron estos crímenes, mientras que \citet{Barreto2019} explora las responsabilidades estatales en casos de desapariciones forzadas. Documentos como los de \citet{CCEEU2012} y los informes del \citet{CNMH2015, CNMH2018} ofrecen un panorama detallado del fenómeno, al igual que los análisis de \citet{HRW2015, HRW2021} sobre el papel de los altos mandos en las ejecuciones. Estudios como los de \citet{CardenasVilla2013} examinan las relaciones entre doctrina militar y ejecuciones extrajudiciales, mientras que \citet{Bonilla2017} analiza los discursos enfrentados en torno a los falsos positivos. Otros trabajos como los de \citet{FOR_CCEEU2014, Giraldo2021, Palencia2011} abordan la dimensión ideológica, simbólica y pedagógica de esta forma de violencia. Finalmente, \citet{CINEP2019} ofrece un balance anual sobre las violaciones de derechos humanos, visibilizando la persistencia de prácticas represivas camufladas bajo lógicas institucionales.

Si bien la mayoría de estos estudios han adoptado enfoques cualitativos —centrados en análisis de caso, reconstrucciones históricas y marcos jurídicos—, son escasos aquellos que incorporan herramientas cuantitativas para el estudio de patrones, correlaciones o estructuras sistemáticas asociadas a estos crímenes. Esta limitación ha dificultado una caracterización más completa de las dinámicas subyacentes a la comisión de los falsos positivos.

En otros contextos, como Argentina o Guatemala, se han desarrollado metodologías mixtas que combinan análisis cualitativo con técnicas estadísticas y minería de datos para identificar patrones de violencia estatal \citep{ball1999state, guzman2015patterns}. Estos enfoques permiten no solo validar hallazgos previos, sino también descubrir nuevas regularidades que suelen pasar desapercibidas en estudios exclusivamente descriptivos. Herramientas como el análisis de redes aplicado a texto se han mostrado particularmente útiles para representar relaciones entre actores, conceptos y eventos a partir de testimonios judiciales, permitiendo una visión estructural y relacional de los discursos presentados ante la JEP \citep{diesner2005exploration, roberts2018structural}.

El uso de técnicas de procesamiento de lenguaje natural (PLN) y aprendizaje automático (ML) amplía aún más las posibilidades de análisis, al facilitar la detección automatizada de patrones narrativos, la identificación de temas emergentes y la evaluación de la carga emocional de los discursos. En otras experiencias internacionales, como las investigaciones de \citet{price2009patterns}, se ha demostrado que estos enfoques híbridos mejoran la documentación de violaciones a los derechos humanos, contribuyen a la formulación de políticas de reparación y refuerzan los mecanismos de prevención. En el contexto colombiano, la aplicación de estas metodologías puede ofrecer una base empírica sólida que complemente el trabajo cualitativo de las organizaciones de memoria y contribuya a los pilares de verdad, justicia, reparación y no repetición. 

\subsection{Justicia transicional}

La justicia transicional en Colombia ha experimentado una evolución significativa desde su formalización, particularmente a partir de la Ley de Justicia y Paz de 2005 y, de forma más amplia, con el Acuerdo Final de Paz de 2016 entre el Estado colombiano y las FARC-EP. Este campo ha sido ampliamente abordado por la literatura académica, con investigaciones centradas en comprender, evaluar y fortalecer los mecanismos creados para afrontar los crímenes cometidos durante el conflicto armado y reparar integralmente a las víctimas. De acuerdo con \citet{AponteGarciaSanchezArteaga2024}, los primeros estudios se enfocaron en la documentación de atrocidades y en la implementación de marcos normativos dirigidos a la reparación, destacando el papel que jugó la Ley de Justicia y Paz en la consolidación de los primeros esfuerzos institucionales en esta materia. Con el tiempo, y especialmente a partir de la firma del Acuerdo de 2016, el enfoque de los estudios se ha ampliado y sofisticado, incorporando dimensiones como la participación activa de las víctimas, el enfoque territorial, el análisis interseccional y la transversalización del enfoque de género \citep{EspinosaDiazRios2022}.

Una de las principales innovaciones introducidas por el Acuerdo fue la creación del Sistema Integral de Verdad, Justicia, Reparación y No Repetición, que articula tres mecanismos interdependientes: la JEP, la Comisión de la Verdad y la Unidad de Búsqueda de Personas dadas por Desaparecidas. Este sistema ha sido valorado positivamente por su capacidad de adaptarse a las condiciones sociopolíticas del país, al integrar principios restaurativos, participación comunitaria y una visión transformadora de la justicia transicional, orientada a modificar las estructuras de exclusión y violencia que dieron origen y continuidad al conflicto armado \citep{VelasquezRuiz2022}.

Los estudios más recientes han adoptado enfoques comparativos e interdisciplinarios, analizando las diferencias entre los procesos de justicia transicional aplicados a distintos actores armados —como las AUC y las FARC-EP— y evaluando el rol de la sociedad civil, así como el impacto de estos procesos en las poblaciones más afectadas por la violencia. Investigaciones como las de \citet{Rettberg2016} y \citet{GarridoOrtolá2023} han examinado la manera en que estos mecanismos han sido recibidos y adaptados por distintas comunidades. Paralelamente, otros trabajos han subrayado los avances y retos en materia de inclusión, visibilizando las experiencias diferenciales de comunidades afrodescendientes, indígenas, mujeres y personas LGBTI$+$, tradicionalmente marginadas de los procesos de justicia ordinaria y transicional \citep{Santamaria2023}.

Dentro del Sistema Integral, la Jurisdicción Especial para la Paz se erige como el componente judicial central, encargado de investigar, juzgar y sancionar las violaciones más graves al Derecho Internacional Humanitario y al Derecho Internacional de los Derechos Humanos. Su mandato comprende tanto crímenes de guerra como de lesa humanidad cometidos antes del 1 de diciembre de 2016, y se orienta por principios restaurativos, buscando esclarecer la verdad, ofrecer justicia a las víctimas y contribuir a la reconciliación nacional. De acuerdo con \citet{AponteGarciaSanchezArteaga2024}, la JEP ha logrado avances sustantivos en la identificación y priorización de macrocasos emblemáticos —como el denominado Caso 03 sobre ejecuciones extrajudiciales— y ha permitido la participación activa de las víctimas en las distintas etapas procesales, garantizando su centralidad en la búsqueda de justicia.

Aunque la JEP ha sido objeto de críticas por su enfoque restaurativo y la posibilidad de imponer sanciones propias diferentes a la prisión, diversos estudios reconocen su carácter innovador dentro del campo de la justicia transicional. En particular, \citet{VelasquezRuiz2022} destaca la incorporación de prácticas de justicia ancestral, mecanismos de participación comunitaria y una comprensión contextualizada del conflicto, lo que ha permitido una aproximación más inclusiva, sensible al territorio y representativa de las voces históricamente silenciadas. Sin embargo, la implementación de la JEP ha enfrentado obstáculos significativos, entre ellos problemas operativos, tensiones institucionales y una fuerte polarización política y mediática que ha condicionado la percepción pública sobre su legitimidad y eficacia \citep{BoteroRojas2023}. Estos desafíos evidencian la necesidad de fortalecer los componentes institucionales y sociales de la justicia transicional, y reafirman el valor de evaluaciones rigurosas que permitan garantizar su sostenibilidad en el largo plazo.

\section{Metodología}

Esta sección describe la metodología empleada para el análisis computacional de los testimonios registrados en los videos del Caso 03 de la JEP, disponibles en su canal oficial de YouTube. La población objetivo está compuesta por la totalidad de los videos emitidos hasta el 2 de febrero de 2025, accesibles en \url{https://www.youtube.com/@jepcolombia}. A partir de técnicas de procesamiento del lenguaje natural (PLN), análisis de sentimiento y modelado de redes semánticas, se plantea una estrategia integral para extraer, organizar y examinar las narrativas de víctimas y comparecientes relacionadas con ejecuciones extrajudiciales. El procedimiento contempla la recopilación y depuración de los datos textuales, su análisis estadístico y lingüístico, así como la detección de patrones discursivos mediante algoritmos de agrupamiento y métricas de centralidad aplicadas a grafos léxicos. Esta aproximación permite realizar una exploración sistemática del contenido, orientada a evaluar la carga emocional de los discursos, identificar tópicos predominantes y comparar las dinámicas narrativas entre distintos subcasos y tipos de actores involucrados en el conflicto armado.

Para extraer la información, se emplearon las bibliotecas de Python \texttt{youtube\_transcript\_api} y \texttt{pytube}, que permiten obtener transcripciones automáticas de los videos del Caso 03 disponibles en el canal oficial de la JEP, siempre que cuenten con subtítulos habilitados. Se analizaron un total de 441 videos. Posteriormente, se clasificó cada testimonio como proveniente de una víctima o de un compareciente, contrastando esta información con diversas fuentes, entre ellas la página institucional de la JEP. Además, se identifican seis subcasos territoriales —Antioquia, Norte de Santander, Meta, Casanare, Costa Caribe y Huila— con base en las listas de reproducción del canal. Para asignar correctamente cada video a su subcaso, se verifican tanto las listas como los títulos, evaluando si contienen referencias explícitas a la región correspondiente, con el fin de asegurar una clasificación precisa.

\begin{figure}[!htb]
    \centering
    \includegraphics[width=1\linewidth]{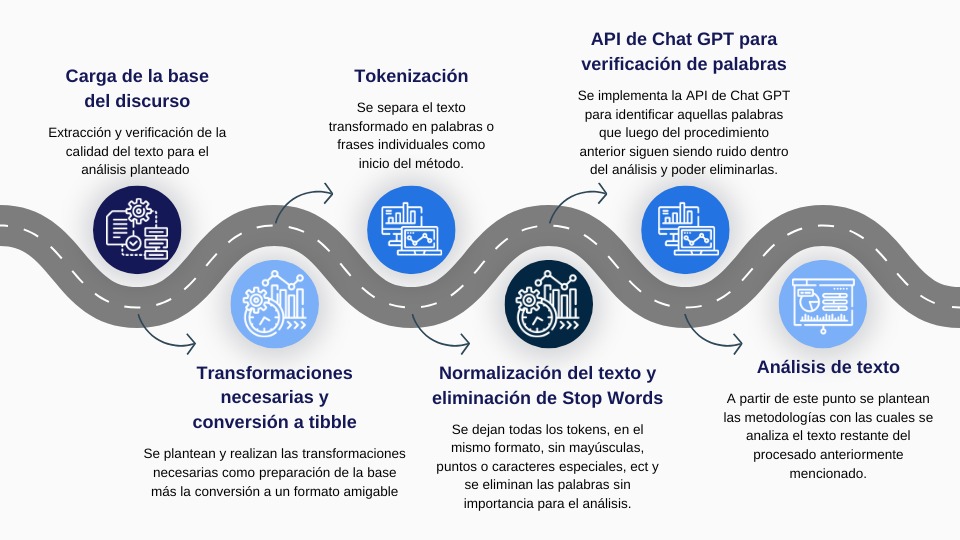}
    \caption{Flujograma del preprocesamiento.}
    \label{fig_flujo}
\end{figure}

El análisis de los videos del Caso 03 se realizó mediante técnicas de Procesamiento del Lenguaje Natural (PLN), una disciplina que integra la lingüística computacional—basada en reglas formales para modelar el lenguaje humano—con modelos estadísticos y algoritmos de \textit{machine learning}. Esta combinación permite a los sistemas computacionales reconocer, interpretar y generar texto y voz, facilitando así el análisis automatizado de grandes volúmenes de datos lingüísticos \citet{chowdhary2020natural}. El procesamiento comenzó con una etapa de preprocesamiento aplicada a los subtítulos de los videos del Caso 03, cuyo procedimiento se ilustra en la Figura \ref{fig_flujo}. Una vez preprocesados los textos, se aplicó un análisis de sentimiento que permitió identificar y extraer opiniones subjetivas, evaluando las emociones, actitudes y polaridades —positiva, negativa o neutra— expresadas en el lenguaje \citet{medhat2014sentiment}.

Para el análisis de sentimiento, se incorporó un paso adicional en el preprocesamiento: la lematización de las palabras. Esta técnica consiste en reducir cada palabra a su forma base o \textit{lema}, lo cual facilita el reconocimiento y agrupamiento de términos morfológicamente relacionados. Al aplicar lematización, se mejora la coherencia semántica del análisis y se reduce la redundancia léxica en el corpus. En este estudio, la lematización se realiza utilizando el modelo \texttt{UDPipe}, una herramienta de procesamiento sintáctico y morfológico desarrollada para múltiples lenguas, que permite llevar a cabo segmentación, etiquetado gramatical (\textit{POS tagging}) y lematización con alta precisión \citet{straka2016udpipe}.

Una vez completado el proceso de lematización, se llevó a cabo un análisis descriptivo preliminar del corpus textual. Este análisis incluye la identificación de las palabras más frecuentes mediante el cálculo de frecuencias absolutas, así como la representación visual del vocabulario dominante a través de una nube de palabras (\textit{word cloud}), lo que permite observar tendencias léxicas generales. Posteriormente, se aprovecharon las capacidades de etiquetado morfosintáctico del modelo \texttt{UDPipe}, se extrajeron de forma específica los verbos y adjetivos más frecuentes presentes en el corpus.
 
Tras el análisis descriptivo preliminar, se asignó a cada lema un puntaje de polaridad utilizando una versión traducida del diccionario léxico \texttt{AFINN} \cite{gale2023enhancements}. Este recurso atribuye a las palabras valores enteros en un rango de -5 a +5, donde los valores negativos indican emociones desfavorables o negativas, y los positivos, emociones favorables o positivas. Esta codificación permite cuantificar la carga emocional de los textos y facilita la agregación de puntajes por segmento o documento \cite{aung2018lexicon}. Dado que algunos lemas no se encuentran contemplados en el diccionario \texttt{AFINN}, se complementó el análisis con el diccionario léxico \texttt{CountWordsFree}, el cual permite ampliar la cobertura semántica al asignar connotaciones positivas o negativas a términos adicionales. Esta estrategia combinada contribuye a una estimación más robusta del sentimiento general del corpus.

Una vez identificadas las palabras con connotaciones positivas y negativas, junto con sus respectivos puntajes, se aplicaron dos pruebas de hipótesis para contrastar la distribución del sentimiento en el corpus. La primera prueba evalúa si el promedio de los puntajes negativos es estadísticamente mayor que el de los positivos; la segunda analiza si la proporción de sentimientos negativos supera a la proporción de positivos. Dado que, en el marco del proceso restaurativo, se considera deseable que los sentimientos positivos predominen o, al menos, se mantengan en equilibrio con los negativos, ambas pruebas se formulan como contrastes unilaterales.

Se empleó la prueba \textit{t} para medias pareadas, una técnica ampliamente utilizada para comparar magnitudes asociadas a categorías opuestas dentro de un mismo conjunto de datos \citet{gosset1908probable}. Para verificar los supuestos de esta prueba, se aplica la prueba de Shapiro-Wilk, que evalúa la normalidad de las diferencias \citet{shapiro1965analysis}. Si el supuesto de normalidad no se cumple, se recurre a la prueba no paramétrica de Wilcoxon para muestras pareadas, una alternativa robusta frente a violaciones de normalidad, frecuentemente empleada en estudios de análisis de sentimiento sobre datos no paramétricos \citet{wilcoxon1945individual}.

Posteriormente, se generó una red de bigramas y \textit{skipgramas} a partir de los subtítulos concatenados de los videos. En el caso de los bigramas, el texto se transforma en unidades formadas por dos palabras consecutivas, mientras que los \textit{skipgramas} permiten incluir palabras intermedias entre los términos relacionados. A partir de estos pares, se construyó un grafo donde los nodos representan palabras y las aristas indican relaciones de consecutividad o proximidad. Este procedimiento puede realizarse mediante dos enfoques: en el primero, se eliminan las palabras vacías (\textit{stop words}), como artículos, pronombres o preposiciones, antes de generar los bigramas o \textit{skipgramas} \citet{silge2017}; en el segundo, se construyen primero los bigramas y luego se eliminan aquellos que contienen palabras vacías. Finalmente, se selecciona el enfoque que logra una mayor agrupación semántica y mejor claridad temática en los resultados.

En este caso particular, se procesan todas las transcripciones eliminando previamente las palabras vacías, y luego se generan los bigramas. Una vez construida la red de bigramas (\textit{skipgramas}), se aplica una metodología para conservar únicamente aquellos más relevantes, dado que muchos presentan frecuencias bajas. Para ello, se calcula el sesgo $s$ de la distribución de frecuencias, una medida de asimetría definida como:
\[
s = \frac{n}{(n-1)(n-2)} \sum_{i=1}^{n} \left( \frac{x_i - \bar{x}}{s} \right)^3
\]
donde \( n \) es el número total de observaciones, \( x_i \) representa cada valor observado, \( \bar{x} \) es la media, y \( s \) es la desviación estándar. Un sesgo positivo indica una cola más pronunciada hacia la derecha; uno negativo, hacia la izquierda \citet{joanes_gill_1998}. El procedimiento consiste en eliminar los bigramas (\textit{skipgramas}) cuya frecuencia sea inferior a un umbral \( v = 1, \dots, m \), siendo \( m \) la frecuencia máxima observada. Luego, se grafica el valor del sesgo frente a estos umbrales y se selecciona aquel punto en que el sesgo se estabiliza o deja de variar significativamente. Este criterio es análogo al método del codo utilizado en el análisis de componentes principales \citet{marutho2018elbow}.

Cabe señalar que este método suele ofrecer resultados satisfactorios en redes extensas, es decir, aquellas con una gran cantidad de videos o datos textuales. En redes pequeñas o poco densas, el umbral de frecuencia se determina empíricamente, ajustándose según los resultados que ofrezcan mayor claridad analítica. A partir de la red final de bigramas (\textit{skipgramas}), se implementaron distintos métodos de partición disponibles en la biblioteca \texttt{igraph} de \texttt{R}. Esta técnica, conocida como detección de comunidades, corresponde a un enfoque no supervisado que permite identificar grupos homogéneos de nodos con base en sus patrones de conexión.

A continuación, se implementan métodos de agrupamiento de vértices con el propósito de identificar conjuntos de términos que comparten características comunes a partir de sus relaciones semánticas en el grafo construido. Los algoritmos de agrupamiento aplicados a grafos generan una partición \( C = \{C_1, \dots, C_K\} \) del conjunto de vértices \( V \) de un grafo \( G = (V, E) \), de manera que el número de aristas que conectan vértices de diferentes grupos \( C_k \) y \( C_\ell \) sea relativamente pequeño en comparación con el número de conexiones internas dentro de cada grupo \( C_k \) \cite{kolaczyk2014}. Para evaluar la calidad de esta partición, se utiliza la medida de \textit{modularidad}, que cuantifica el grado de separación entre los grupos generados. La modularidad se define como:
\[
\text{mod}(C) = \frac{1}{2m} \sum_{i,j; i \neq j} \left( y_{i,j} - \frac{1}{2m} d_i d_j \right) \delta_{c_i, c_j}
\]
donde \( \mathbf{Y} = [y_{i,j}] \) representa la matriz de adyacencia del grafo, \( m \) es el número total de aristas, \( d_i \) es el grado del vértice \( i \), \( c_i \) indica el grupo al que pertenece el vértice \( i \), y \( \delta_{x,y} \) es la función de Kronecker, que toma el valor 1 si \( x = y \) y 0 en caso contrario \cite{arXiv:1505.03481}.

Una vez ejecutados todos los algoritmos de partición disponibles en la biblioteca \texttt{igraph}, con excepción del algoritmo \texttt{optimal community structure} —recomendado únicamente para redes de pequeño tamaño por su elevada complejidad computacional \cite{igraph_cluster_optimal}—, se selecciona aquel que alcanza el mayor valor de modularidad. Esta métrica permite evaluar la calidad de la partición, ya que refleja el grado de cohesión interna dentro de los grupos frente a su separación respecto a otros. Valores elevados de modularidad indican que la partición identificada por los algoritmos revela una estructura latente no trivial en la red, más allá de lo esperable bajo un modelo de asignación aleatoria de aristas \cite{kolaczyk2014}.

Para interpretar de manera más precisa los grupos de palabras identificados en la red, se emplean técnicas de inteligencia artificial orientadas a inferir su significado y así detectar los tópicos predominantes en las descripciones. Dado que en toda red existen nodos con mayor relevancia estructural que otros, se asigna un orden de importancia descendente a los nodos para facilitar la identificación de términos clave. Esta relevancia se determina mediante la centralidad propia (\textit{eigenvector centrality}), un indicador que define a un nodo como importante si está conectado con otros nodos que también poseen alta centralidad \cite{bonacich2007}. Formalmente, la centralidad propia de un vértice \( v \) se expresa como:
\[
c_{\textsf{E}}(v) = \alpha \sum_{\{u,v\} \in E} c(u)
\]
donde \( c = (c(1), \dots, c(n)) \) representa el vector de centralidades que satisface la ecuación de valores propios \( Yc = \alpha^{-1} c \); aquí, \( Y \) es la matriz de adyacencia del grafo, \( \alpha^{-1} \) es el valor propio dominante de \( Y \), y \( c \) es el vector propio correspondiente \cite{jalili2017}. Siguiendo la convención establecida, se reportan los valores absolutos de las entradas de \( c \) \cite{newman2018}.

Además, al representar gráficamente las redes generadas a partir de los bigramas y \textit{skipgramas} del caso general, los nodos se dimensionan en proporción a su grado de intermediación. Esta medida de centralidad resulta especialmente útil en visualizaciones, ya que permite destacar los nodos que actúan como puentes entre distintas partes de la red, facilitando así la interpretación de su importancia estructural. El grado de intermediación se define formalmente como:
\[
c_{\textsf{B}}(v) = \sum_{s,t \in V; \, s \neq t \neq v} \frac{\sigma(s,t \mid v)}{\sigma(s,t)}
\]
donde \(\sigma(s,t \mid v)\) representa el número de caminos más cortos entre los nodos \(s\) y \(t\) que pasan por \(v\), y \(\sigma(s,t)\) es el número total de caminos más cortos entre \(s\) y \(t\), independientemente de si pasan por \(v\) o no \cite{newman2018}.

Dado que algunas redes analizadas, como la correspondiente al conjunto completo de videos, son especialmente densas, se incorpora el algoritmo \textit{k-core} con el objetivo de mejorar la calidad visual y analítica de las representaciones gráficas. Este algoritmo, perteneciente a la teoría de grafos, permite identificar subestructuras altamente conectadas dentro de la red, y se define como el subgrafo máximo en el cual cada nodo tiene al menos \(k\) conexiones con otros nodos del mismo subgrafo \cite{kong2019}. Para su implementación, se calcula el valor de \(k\) para cada nodo y se grafican únicamente aquellos con valores inferiores a la mediana, con el fin de reducir la sobrecarga visual causada por nodos excesivamente conectados, lo cual permite una mejor interpretación de la estructura semántica subyacente.

Todo el procedimiento descrito se replica de forma sistemática para cada uno de los subcasos territoriales definidos en el Caso 03, así como para las particiones construidas en función del rol de los actores (víctimas y comparecientes). Esta segmentación posibilita un análisis comparativo entre distintos grupos de interés, tanto en un nivel agregado como específico, manteniendo la coherencia metodológica en el análisis de sentimiento, las pruebas de hipótesis y los criterios estadísticos aplicados. Con ello, se busca identificar patrones diferenciados en las narrativas y cargas emocionales de los testimonios, ofreciendo evidencia empírica que complemente el enfoque cualitativo del proceso judicial y restaurativo.

\section{Resultados}

Una vez realizada la categorización de los videos por subcaso y rol de los comparecientes, se obtiene la distribución que se presenta en la Tabla \ref{tab:victimas_victimarios}.

\begin{table}[!htb]
    \centering
    \begin{tabular}{lccc}
    \hline
    Región & Total & Comparecientes & Víctimas \\
    \hline
    Norte de Santander & 14 & 1 & 0 \\
    Costa Caribe & 21 & 10 & 5 \\
    Casanare & 101 & 83 & 16 \\
    Meta & 4 & 0 & 1 \\
    Huila & 181 & 136 & 40 \\
    Antioquia & 9 & 2 & 5 \\
    General & 441 & 328 & 77 \\
    \hline
    \end{tabular}
    \caption{Número de videos por subcaso.}
    \label{tab:victimas_victimarios}
\end{table}

Como se observa, los subcasos de Meta y Norte de Santander cuentan con un número muy reducido de videos categorizados por rol —uno cada uno—, lo que dificulta la posibilidad de realizar comparaciones significativas entre víctimas y comparecientes. De manera similar, el subcaso de Antioquia presenta únicamente dos videos correspondientes a comparecientes. Por tanto, se excluyen estos tres subcasos de los análisis diferenciados por rol, con el objetivo de preservar la consistencia metodológica y asegurar la representatividad de los resultados.

\subsection{Análisis de sentimientos}

\begin{figure}[!htb]
    \centering
    \begin{minipage}{0.45\linewidth}
        \centering
        \includegraphics[width=\linewidth]{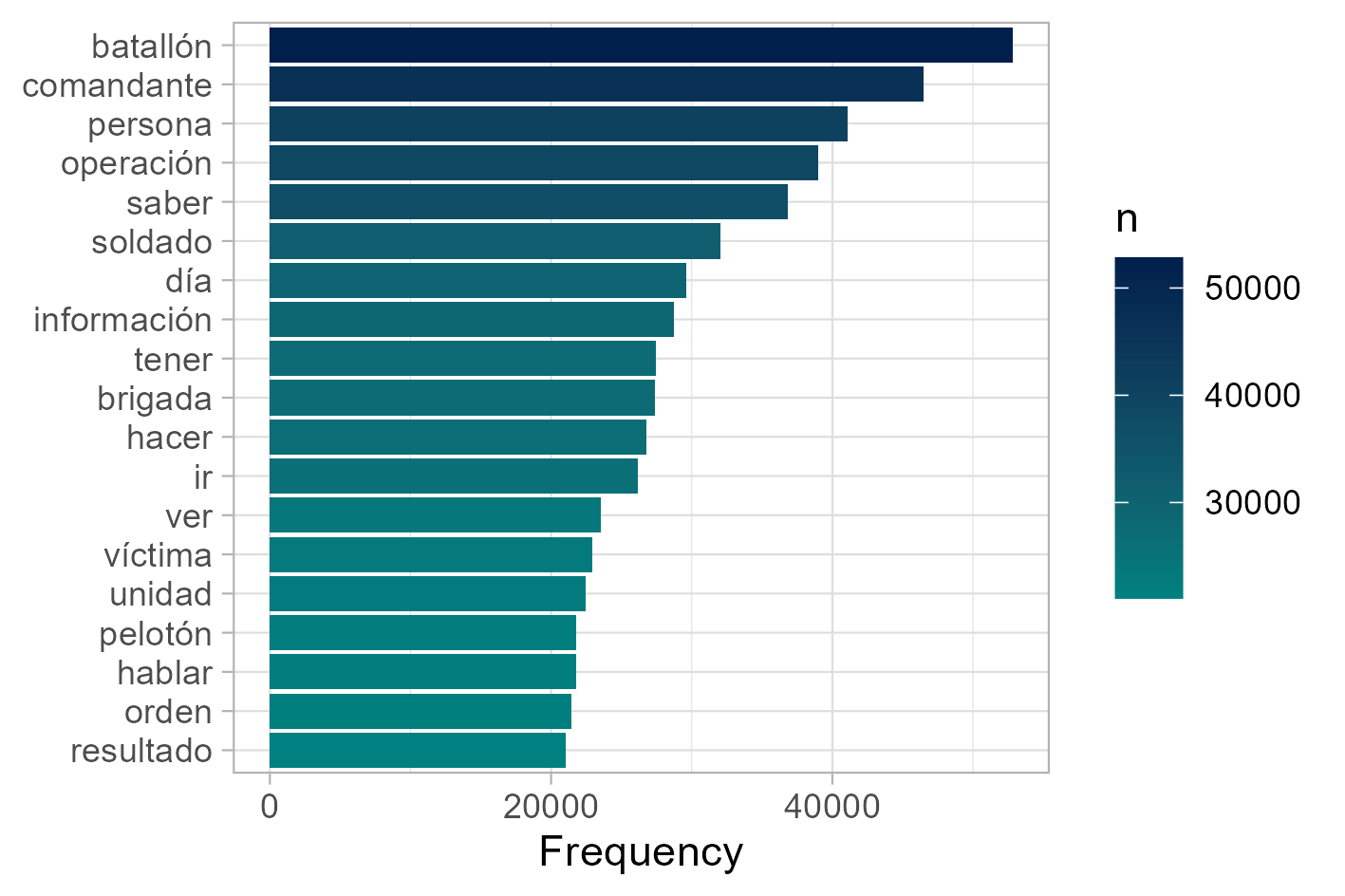}
        \caption{Análisis general: palabras más frecuentes.}
        \label{fig_general_frecuencias}
    \end{minipage}%
    \hfill
    \begin{minipage}{0.45\linewidth}
        \centering
        \includegraphics[width=\linewidth]{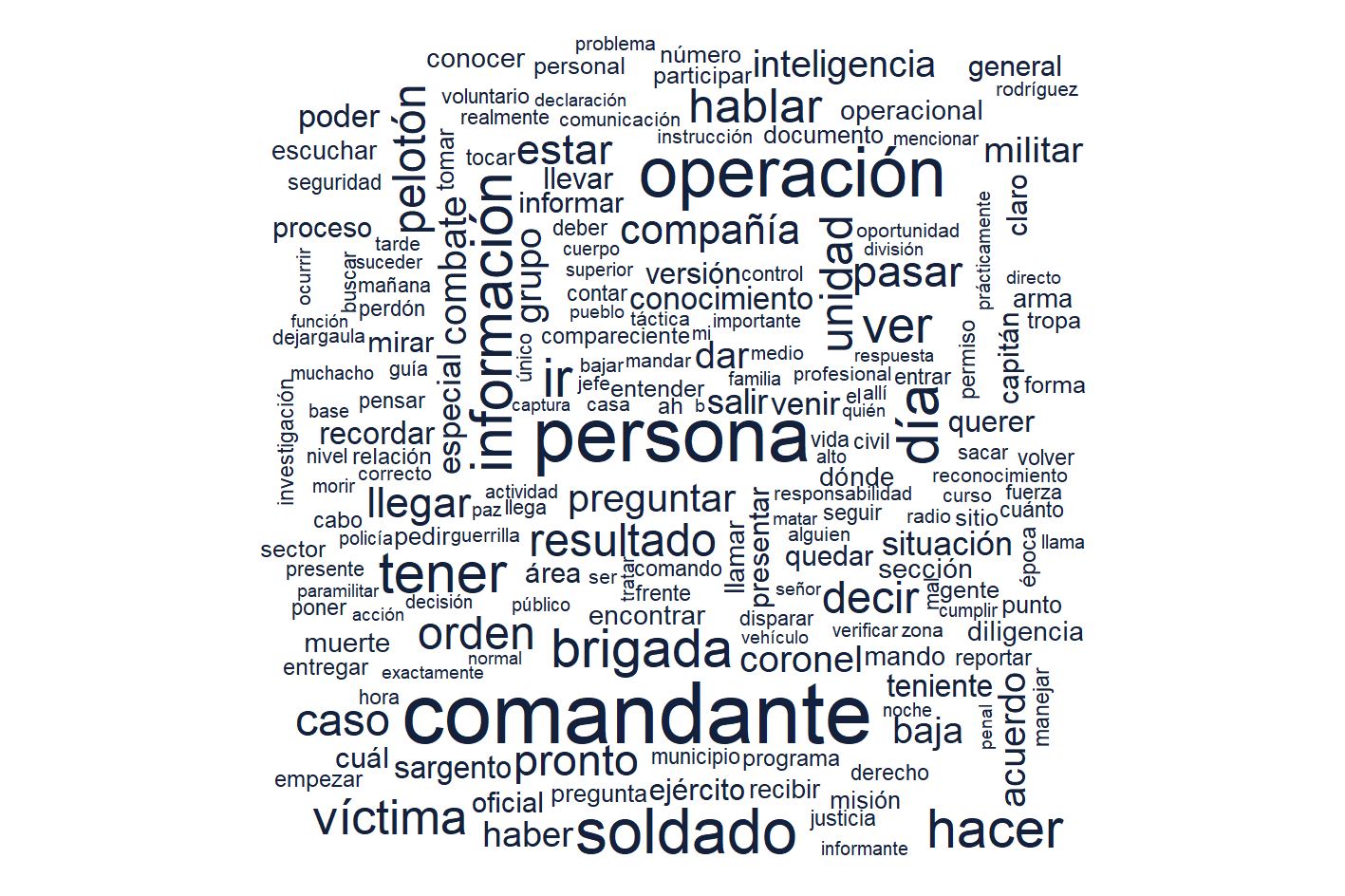}
        \caption{Análisis general: nube de Palabras.}
        \label{fig_general_nube}
    \end{minipage}
\end{figure}

Se realizó un análisis descriptivo del corpus general, cuyos resultados se presentan en la Figura \ref{fig_general_frecuencias} y la Figura \ref{fig_general_nube}. La nube de palabras y el gráfico de frecuencias revelan la alta presencia de términos directamente vinculados al ámbito militar y operativo. Palabras como “batallón”, “comandante”, “soldado”, “persona” y “brigada” remiten a estructuras jerárquicas y unidades propias de las Fuerzas Armadas, mientras que expresiones como “operación”, “día”, “saber” y “hacer” sugieren una narrativa centrada en la ejecución de actividades tácticas y aspectos rutinarios de la vida en servicio. Esta configuración léxica refuerza la centralidad del lenguaje castrense en los testimonios, especialmente por parte de los comparecientes.

Al mismo tiempo, emergen términos que parecen reflejar la voz de las víctimas y otros aspectos del conflicto, como “víctima”, “ver”, “información” y “orden”, lo cual indica la coexistencia de relatos orientados tanto a la exposición de hechos como a la búsqueda de verdad. Adicionalmente, se destacan palabras como “hablar”, “decir” y “recordar”, asociadas al ejercicio testimonial, lo que sugiere una dimensión narrativa orientada a reconstruir los eventos a través de la memoria. En conjunto, estos resultados revelan la superposición de registros discursivos: uno de carácter institucional y operativo, y otro centrado en la experiencia humana y la memoria, lo que refuerza la complejidad del contenido abordado en las audiencias del Caso 03.

\begin{figure}[!htb]
    \centering
    \begin{minipage}{0.35\linewidth}
        \centering
        \includegraphics[width=\linewidth]{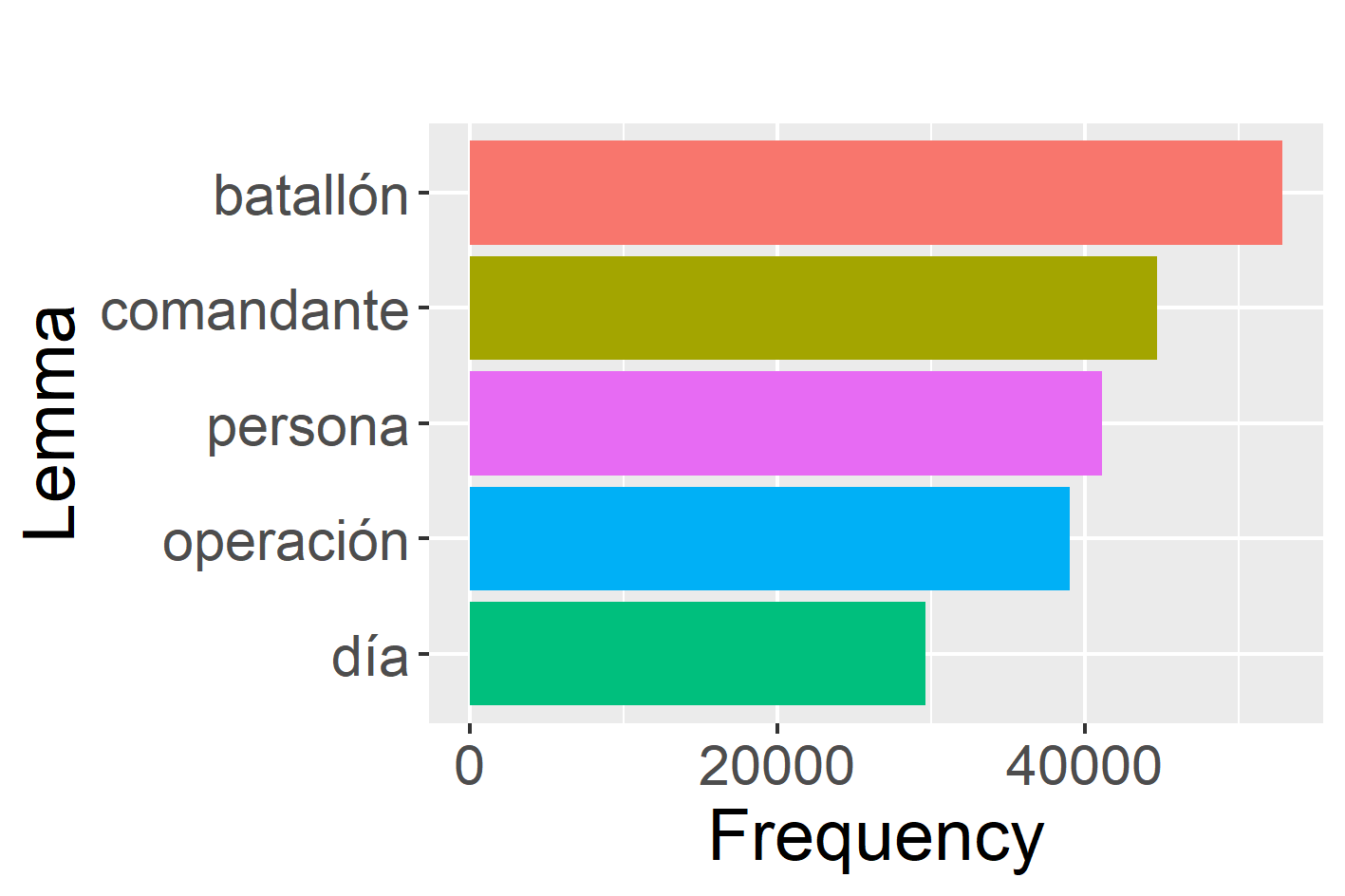}
        \caption{Análisis general: sustantivos más frecuentes.}
        \label{fig_general_sustantivos}
    \end{minipage}%
    \hfill
    \begin{minipage}{0.35\linewidth}
        \centering
        \includegraphics[width=\linewidth]{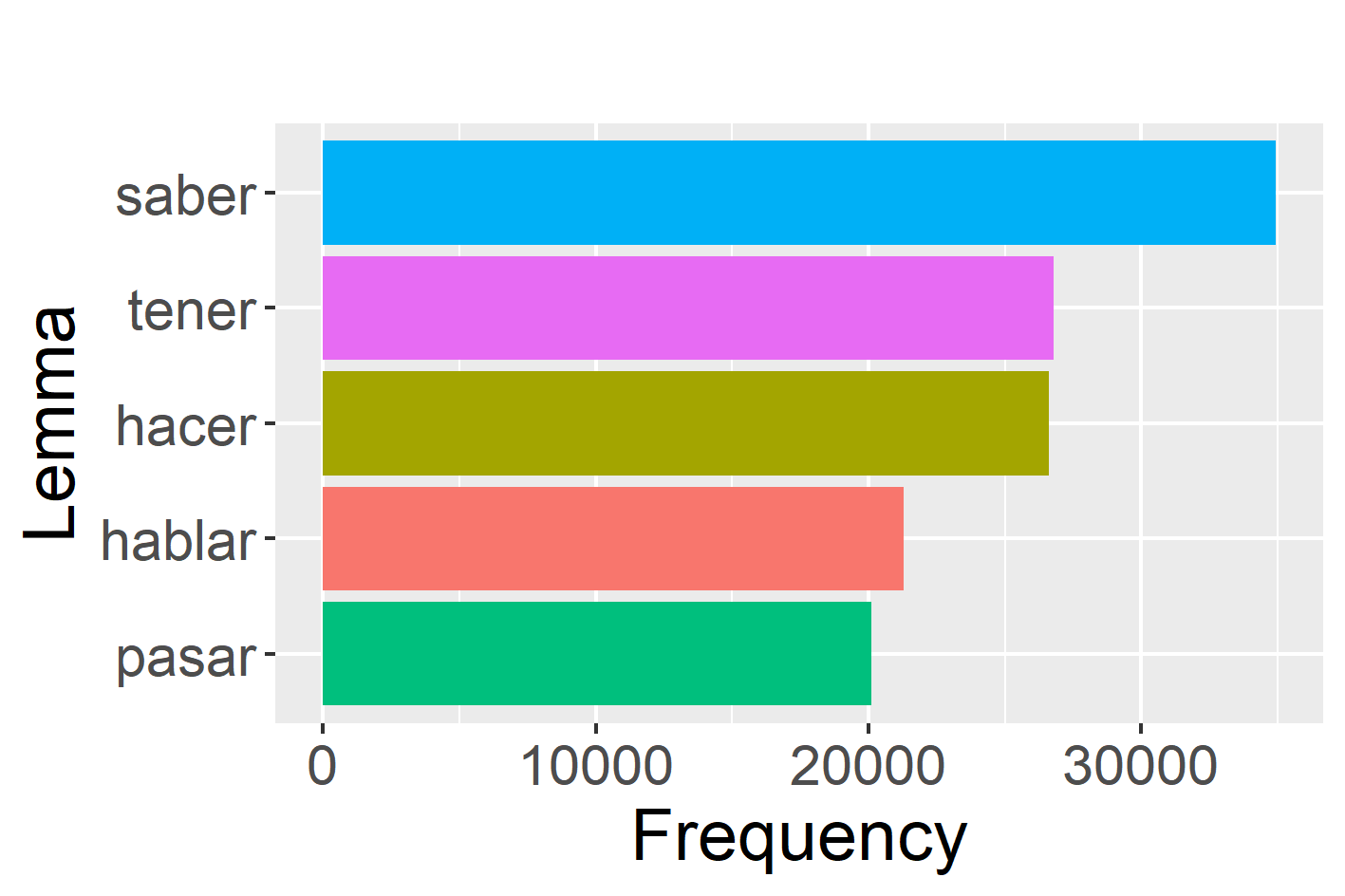}
        \caption{Análisis general: verbos más frecuentes.}
        \label{fig_general_verbos}
    \end{minipage}
\end{figure}

Las Figuras \ref{fig_general_sustantivos} y \ref{fig_general_verbos} presentan los sustantivos y verbos más frecuentes identificados en el análisis del corpus general del caso. En la Figura \ref{fig_general_sustantivos}, los sustantivos predominantes —como “batallón”, “comandante”, “persona”, “operación” y “día”— reflejan una narrativa fuertemente anclada en jerarquías militares, estructuras organizativas y dinámicas operacionales. Estos términos no solo evidencian la centralidad del aparato castrense en los testimonios, sino que también apuntan a un lenguaje institucionalizado que estructura la forma en que se relatan los hechos, priorizando categorías propias del discurso militar.

Por su parte, la Figura \ref{fig_general_verbos} muestra los verbos más frecuentes, entre ellos “saber”, “tener”, “hacer”, “hablar” y “pasar”. Estas acciones verbales ofrecen una ventana hacia los procesos cognitivos y operativos de los actores involucrados. Verbos como “saber” y “tener” sugieren la gestión del conocimiento y la posesión de información clave, mientras que “hacer” y “pasar” remiten a la ejecución de acciones o eventos concretos. El verbo “hablar” destaca por su relación con la dimensión testimonial del discurso, lo que sugiere un espacio de interacción narrativa donde la oralidad adquiere protagonismo. En conjunto, estos patrones léxicos revelan un equilibrio entre la lógica operativa del conflicto armado y la dimensión comunicativa propia de los ejercicios de memoria, verdad y reconocimiento.

\begin{figure}[!htb]
    \centering
    \begin{minipage}{0.48\linewidth}
        \centering
        \includegraphics[width=\linewidth]{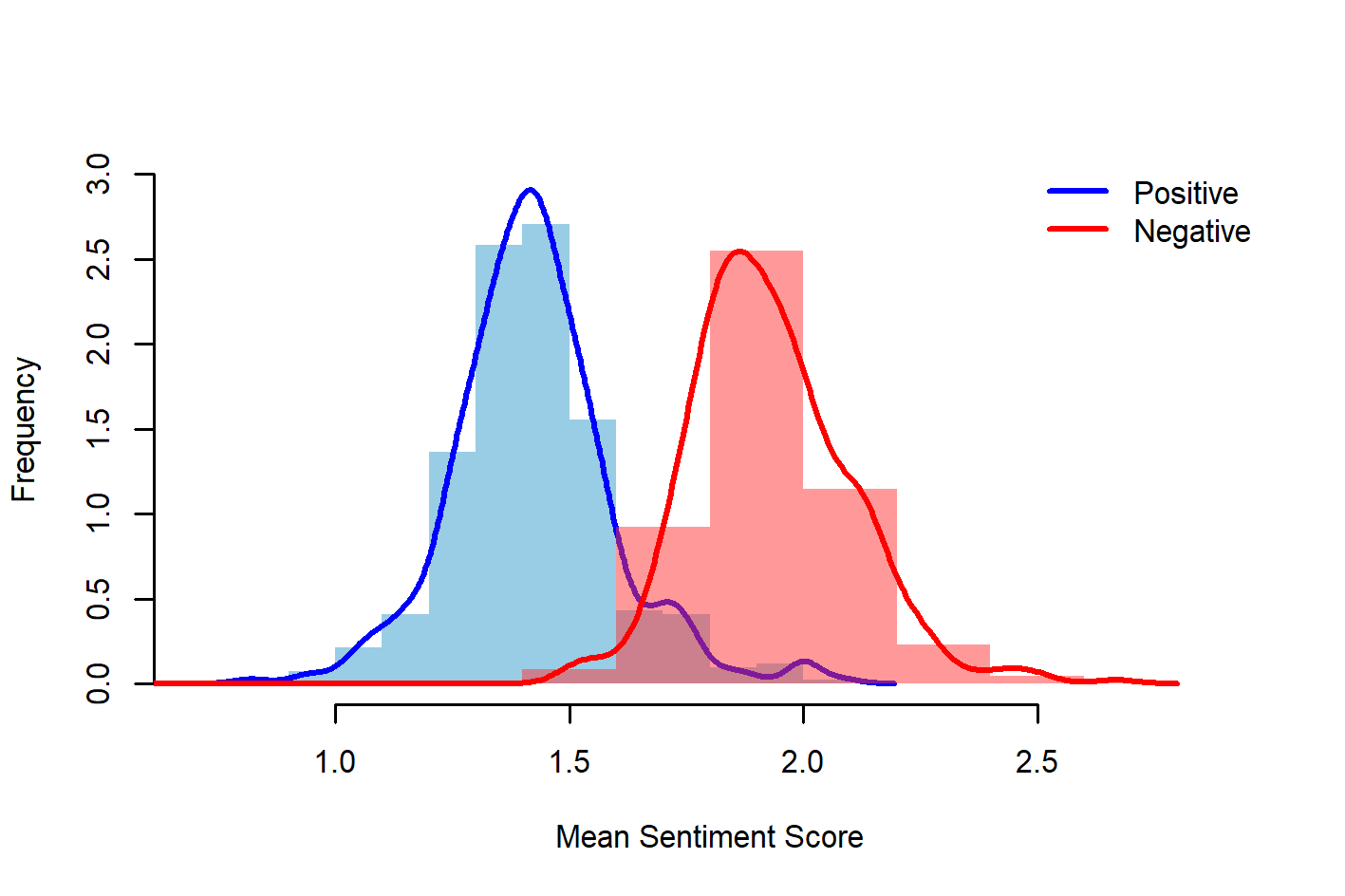}
        \caption{Análisis general: histograma de la distribución media de los sentimientos.}
        \label{fig_general_histograma_sentimientos}
    \end{minipage}%
    \hfill
    \begin{minipage}{0.48\linewidth}
        \centering
        \includegraphics[width=\linewidth]{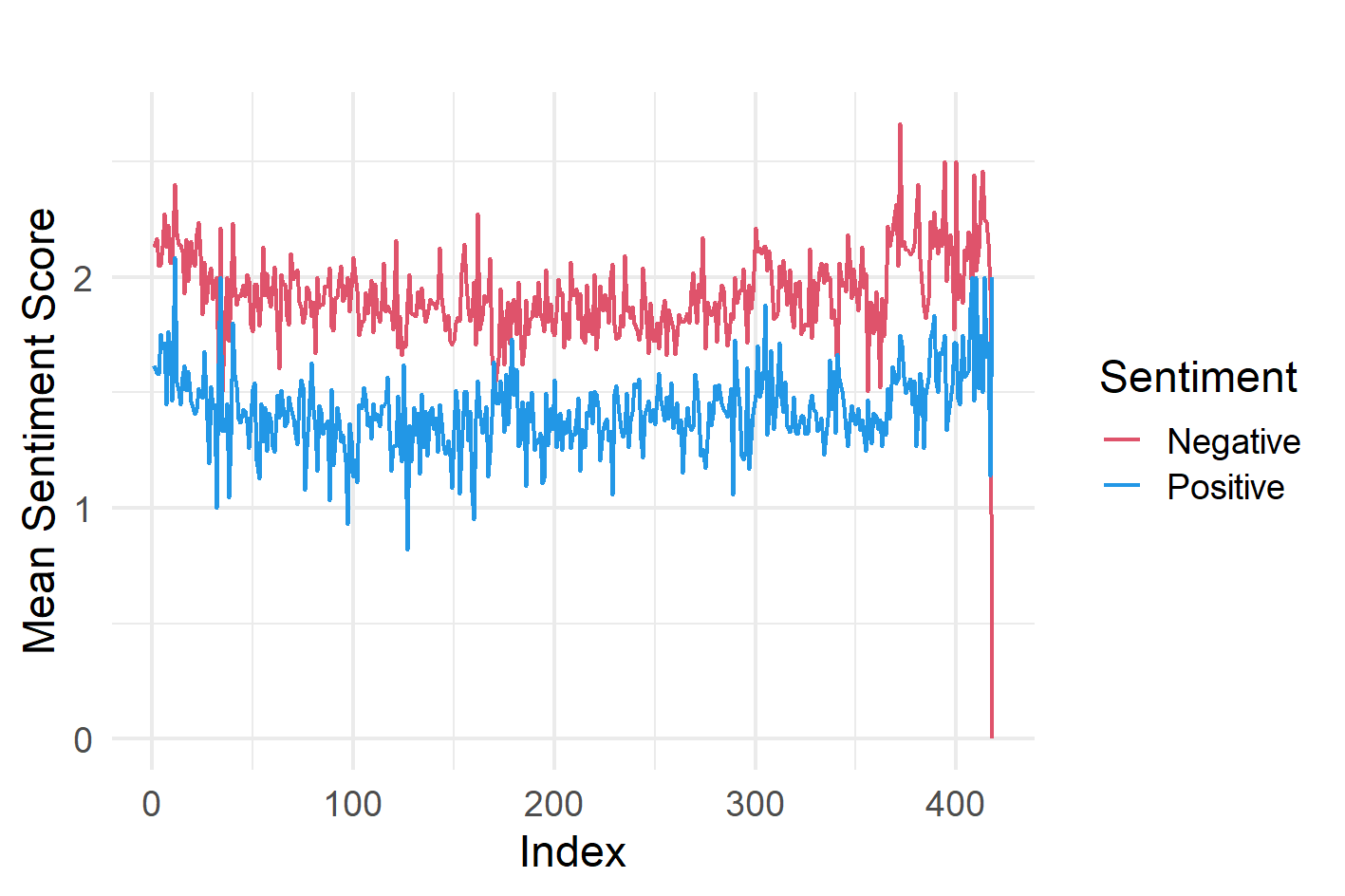}
        \caption{Análisis general: puntaje secuencial de sentimientos positivos y negativos.}
        \label{fig_general_puntaje_secuencial_sentimientos}
    \end{minipage}
\end{figure}

El análisis de los puntajes promedio de sentimiento en los videos del Caso 03 de la JEP revela una predominancia sistemática de sentimientos negativos sobre los positivos a lo largo de toda la discursividad del macrocaso (ver Figuras \ref{fig_general_histograma_sentimientos} y \ref{fig_general_puntaje_secuencial_sentimientos}). En términos cuantitativos, la media de los puntajes negativos alcanza un valor de 1.924, mientras que la media de los positivos es de 1.420. De forma consistente, la mediana de los sentimientos negativos (1.909) también supera a la de los positivos (1.415), lo que refuerza la existencia de una carga emocional adversa sostenida en el discurso general. Estas diferencias no solo reflejan una mayor intensidad de los sentimientos negativos, sino que también sugieren un tono predominantemente doloroso, crítico o reflexivo en los testimonios, posiblemente vinculado a experiencias traumáticas, actos de violencia estatal y demandas de verdad y reparación.

Con el fin de evaluar la significancia estadística de esta diferencia, se aplica la prueba de Shapiro-Wilk al vector de diferencias entre puntajes positivos y negativos, obteniéndose un valor $p < 0.001$, lo que indica que los datos no siguen una distribución normal. Ante la violación del supuesto de normalidad, se recurre a la prueba no paramétrica de Wilcoxon para muestras pareadas, la cual también arroja un valor $p < 0.001$. En consecuencia, se rechaza la hipótesis nula de igualdad de medias, concluyéndose que los sentimientos negativos son significativamente más intensos que los positivos a lo largo del Caso 03. Esta evidencia empírica no solo valida cuantitativamente la prevalencia del dolor y la denuncia en los relatos, sino que también aporta una dimensión emocional clave para comprender la profundidad del daño expresado en el proceso judicial y restaurativo.

\begin{figure}[!htb]
    \centering
    \includegraphics[width=0.65\linewidth]{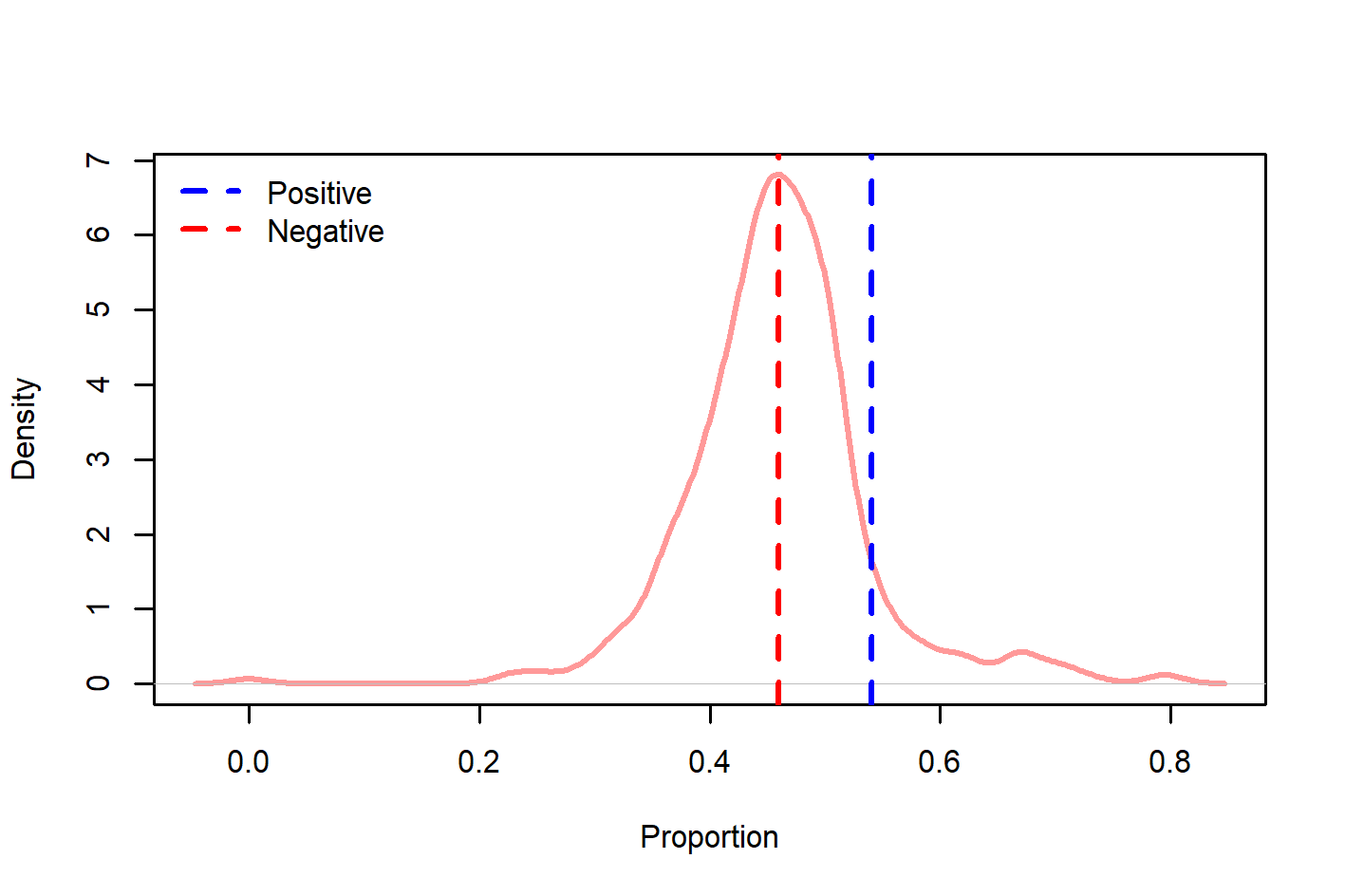}
    \caption{Análisis general: proporción de sentimientos.}
    \label{fig_general_proporcion_sentimientos}
\end{figure}

El análisis de la proporción de palabras con connotaciones positivas y negativas en cada video (ver Figura \ref{fig_general_proporcion_sentimientos}) muestra que, en promedio, las palabras con carga positiva aparecen con mayor frecuencia que aquellas con carga negativa. Esta observación sugiere que, en el discurso global del Caso 03 de la JEP, existe una tendencia general a utilizar un lenguaje que, en términos cuantitativos, incorpora más vocabulario con connotaciones favorables o constructivas. Sin embargo, esta mayor frecuencia de términos positivos contrasta con los hallazgos previos sobre la intensidad del sentimiento, donde se evidenció que las palabras negativas tienen un peso emocional significativamente más alto. Esta aparente discrepancia pone de relieve una dimensión importante del análisis: aunque el lenguaje pueda estar salpicado de expresiones positivas—posiblemente asociadas a la búsqueda de reconciliación, reconocimiento o esperanza—los contenidos más emocionalmente cargados tienden a expresar dolor, denuncia o sufrimiento. En este sentido, la combinación de frecuencia y polaridad revela que el proceso de justicia transicional no solo se articula en torno a la memoria del daño, sino también a través de narrativas que intentan resignificar la experiencia y construir un horizonte de reparación y no repetición.

Para validar la observación sobre la mayor frecuencia de palabras con connotación positiva en el Caso 03 de la JEP, se realizaron pruebas de hipótesis sobre las proporciones de términos positivos y negativos por video. Los resultados descriptivos muestran que la media de las proporciones de palabras positivas es de 0.540, mientras que la de las negativas es de 0.460. De forma consistente, la mediana de las proporciones positivas alcanza un valor de 0.542, superando la mediana de las proporciones negativas, que se sitúa en 0.468. Estos resultados confirman que, en términos de frecuencia relativa, el lenguaje empleado tiende a estar compuesto en mayor proporción por palabras con carga positiva.

No obstante, como se discutió previamente, esta mayor frecuencia no se traduce necesariamente en un sentimiento general más favorable, ya que los sentimientos negativos, aunque menos frecuentes, son más intensos. Esta tensión entre cantidad e intensidad emocional sugiere la coexistencia de dos planos discursivos: por un lado, un lenguaje de reconciliación, reconocimiento o institucionalidad, y por otro, la persistencia del dolor, la denuncia y la memoria del agravio.

Para corroborar estadísticamente esta diferencia en proporciones, se aplicó la prueba de Shapiro-Wilk sobre el vector de diferencias, obteniéndose un valor $p < 0.001$, lo que indica que los datos no se ajustan a una distribución normal. Ante la violación de este supuesto, se utilizó la prueba no paramétrica de Wilcoxon, cuyo resultado fue un valor $p = 1$. Esto implica que no se rechaza la hipótesis nula, por lo que se concluye que, en promedio, los videos contienen una proporción de palabras positivas mayor o igual a la de palabras negativas. Este hallazgo refuerza la idea de una discursividad ambivalente en el macro Caso 03, donde la estructura del lenguaje puede estar orientada a la reparación y al reconocimiento, pero no elimina la persistencia de una carga emocional profundamente negativa asociada al relato de las víctimas y a las responsabilidades asumidas por los comparecientes.

\begin{figure}[!htb]
    \centering
    \begin{minipage}{0.45\linewidth}
        \centering
        \includegraphics[width=\linewidth]{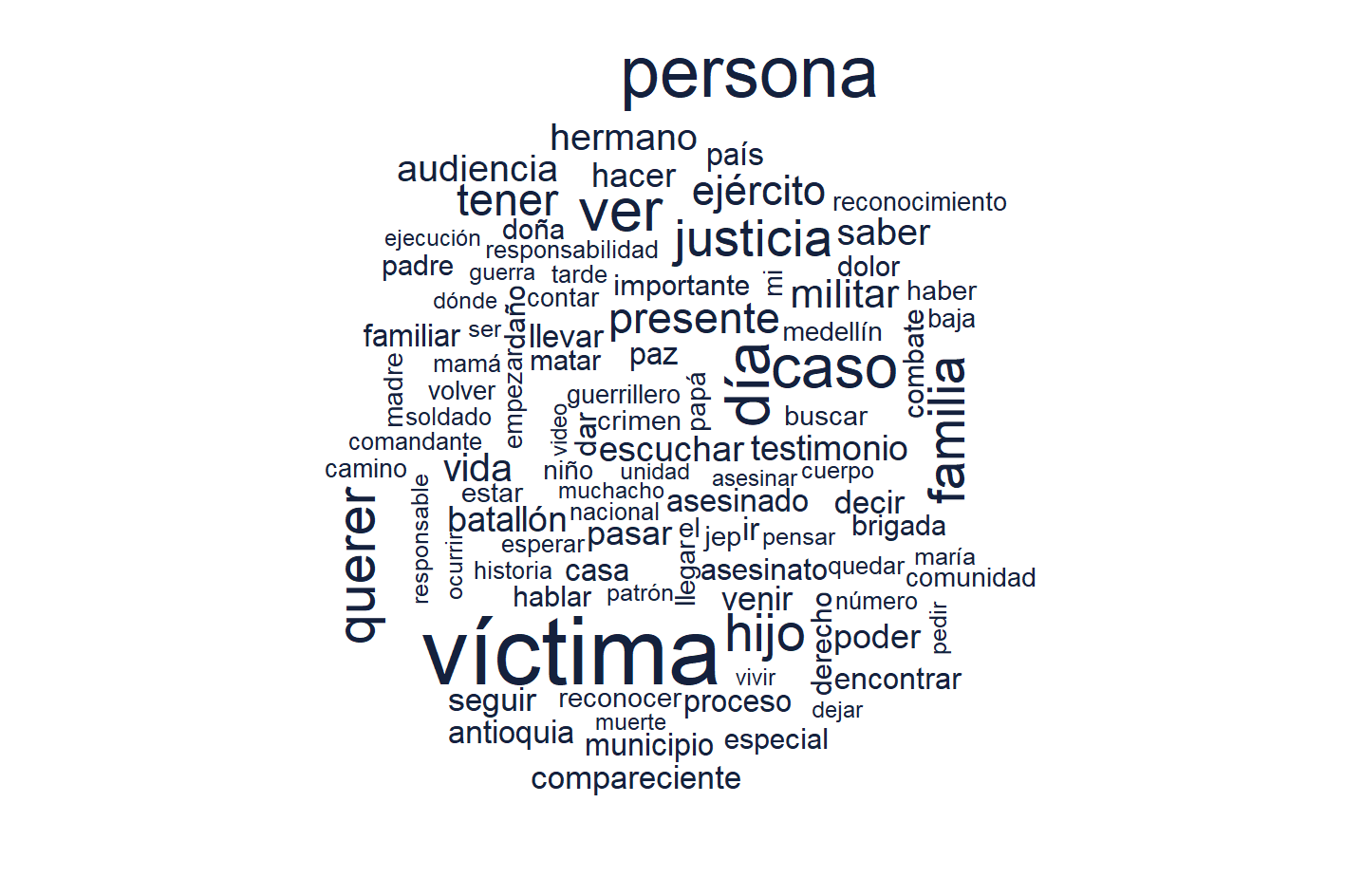}
        \caption{Antioquia: nube de palabras.}
        \label{fig_nube_antioquia}
    \end{minipage}%
    \hfill
    \begin{minipage}{0.45\linewidth}
        \centering
        \includegraphics[width=\linewidth]{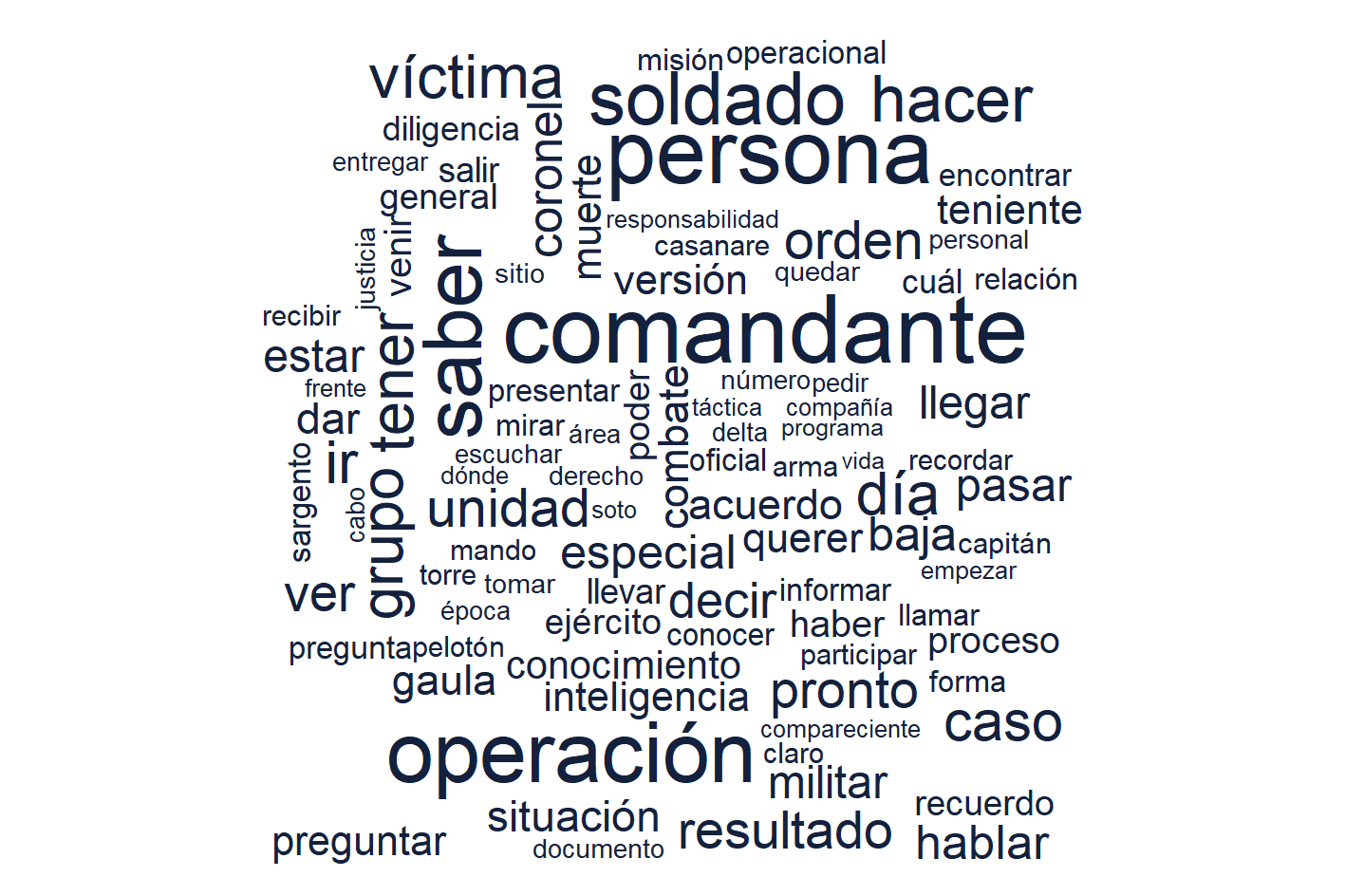}
        \caption{Casanare: nube de palabras.}
        \label{fig_nube_casanare}
    \end{minipage}
\end{figure}

\begin{figure}[H] 
    \begin{minipage}{0.45\linewidth}
        \centering
        \includegraphics[width=\linewidth]{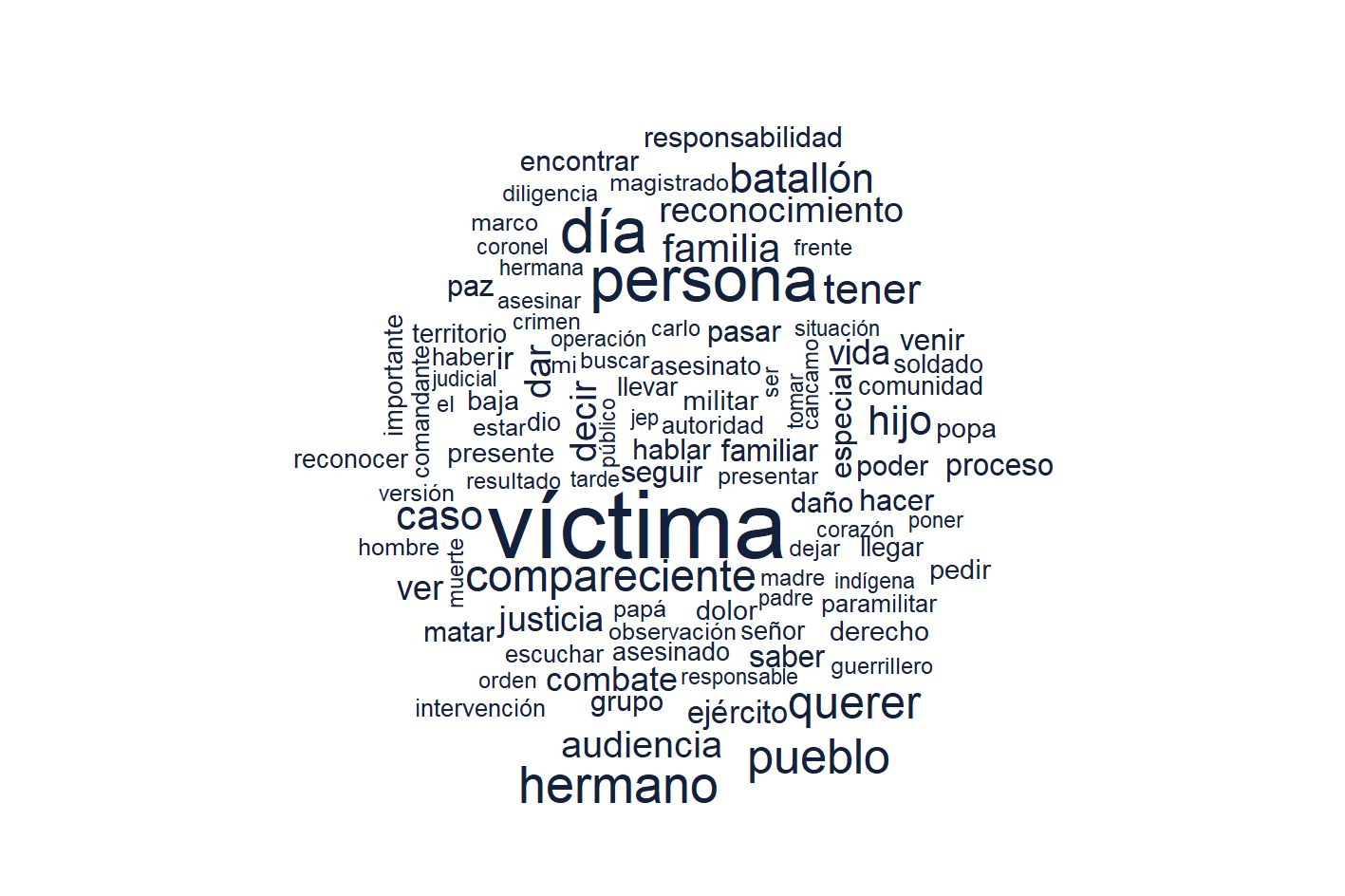}
        \caption{Costa Caribe: nube de palabras.}
        \label{fig_nube_costa_caribe}
    \end{minipage}%
    \hfill
    \begin{minipage}{0.45\linewidth}
        \centering
        \includegraphics[width=\linewidth]{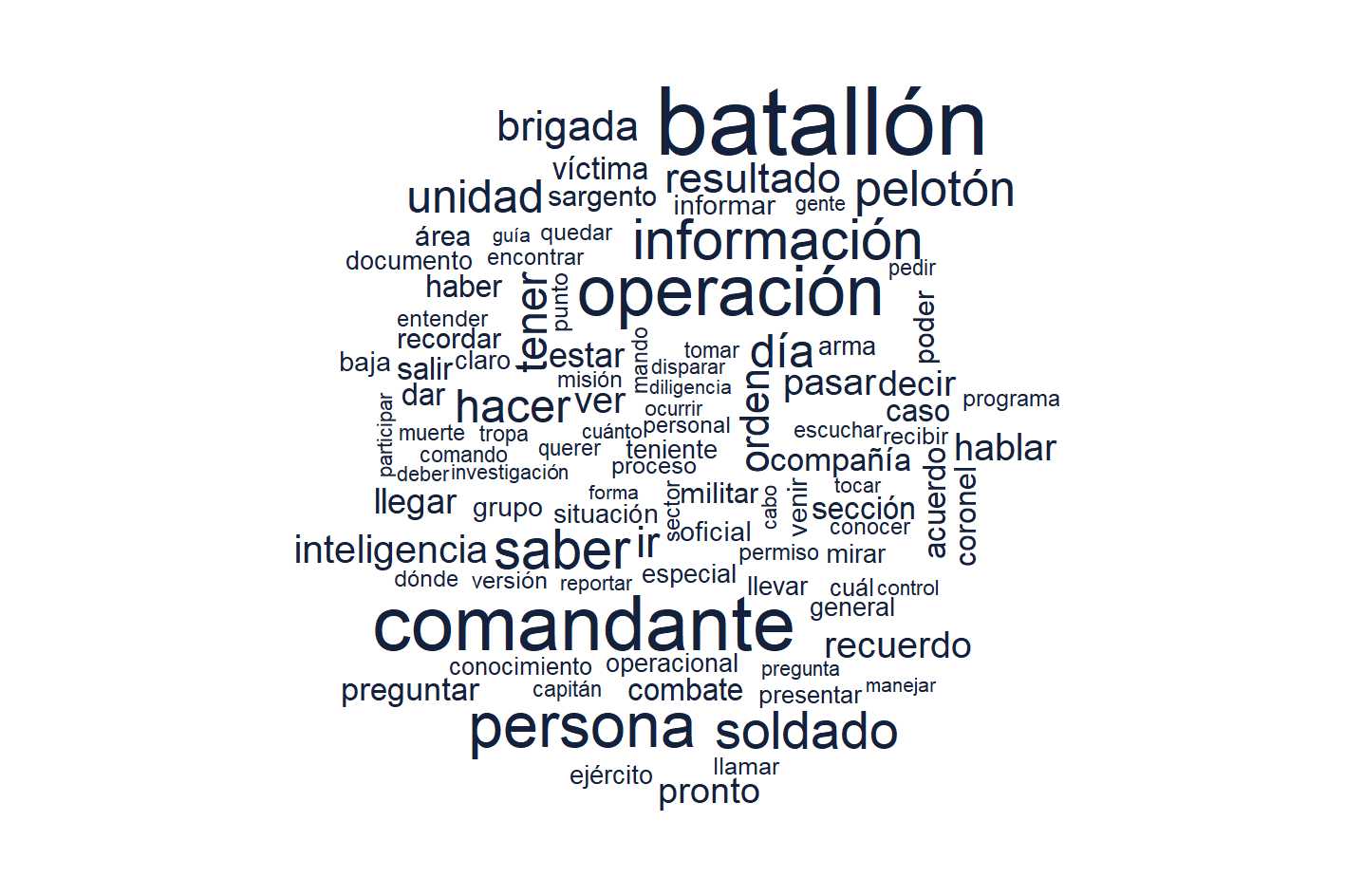}
        \caption{Huila: nube de palabras.}
        \label{fig_nube_huila}
    \end{minipage}
\end{figure}

\begin{figure}[H]    
    \begin{minipage}{0.45\linewidth}
        \centering
        \includegraphics[width=\linewidth]{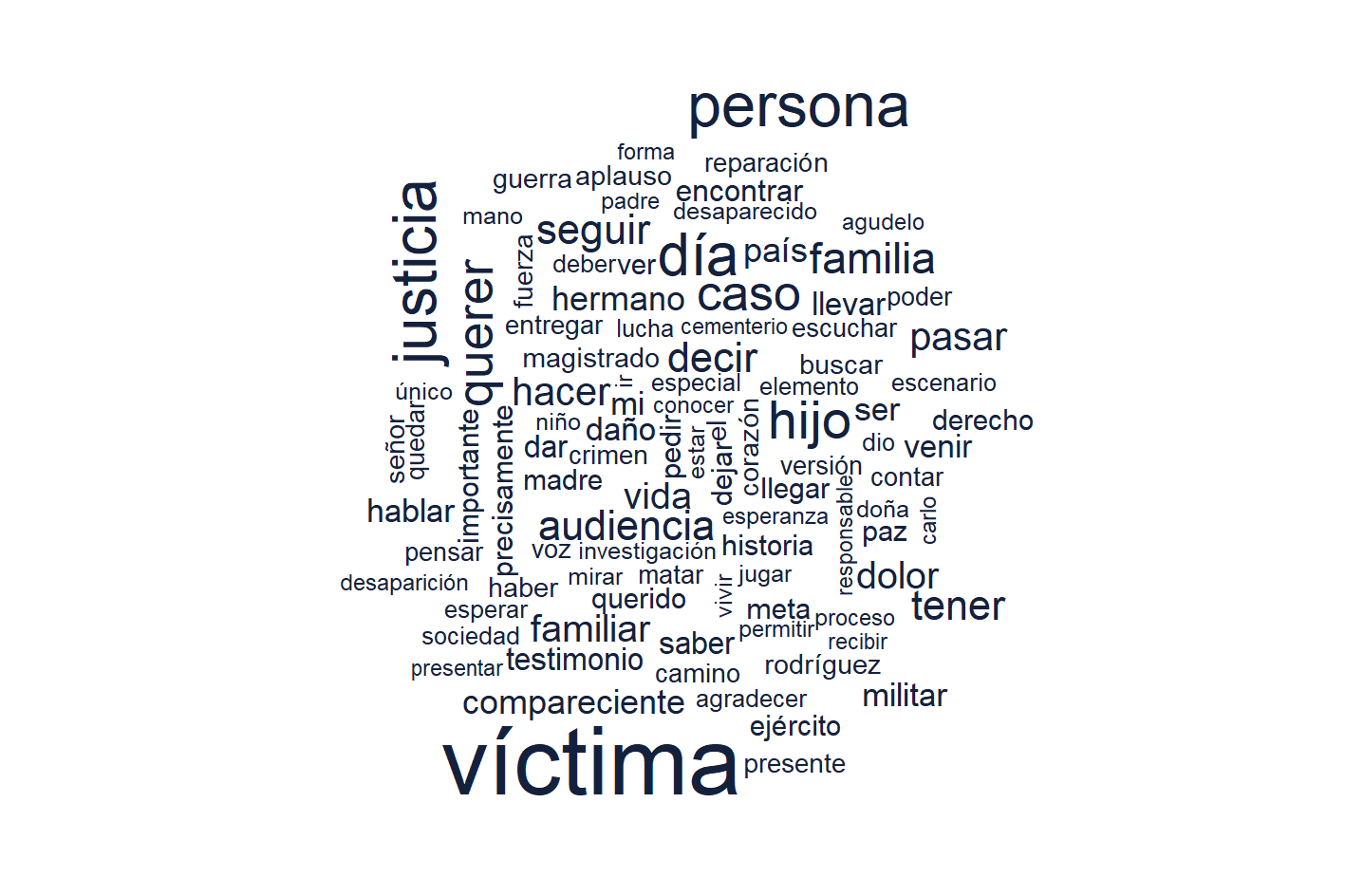}
        \caption{Meta: nube de palabras.}
        \label{fig_nube_meta}
    \end{minipage}%
    \hfill
    \begin{minipage}{0.45\linewidth}
        \centering
        \includegraphics[width=\linewidth]{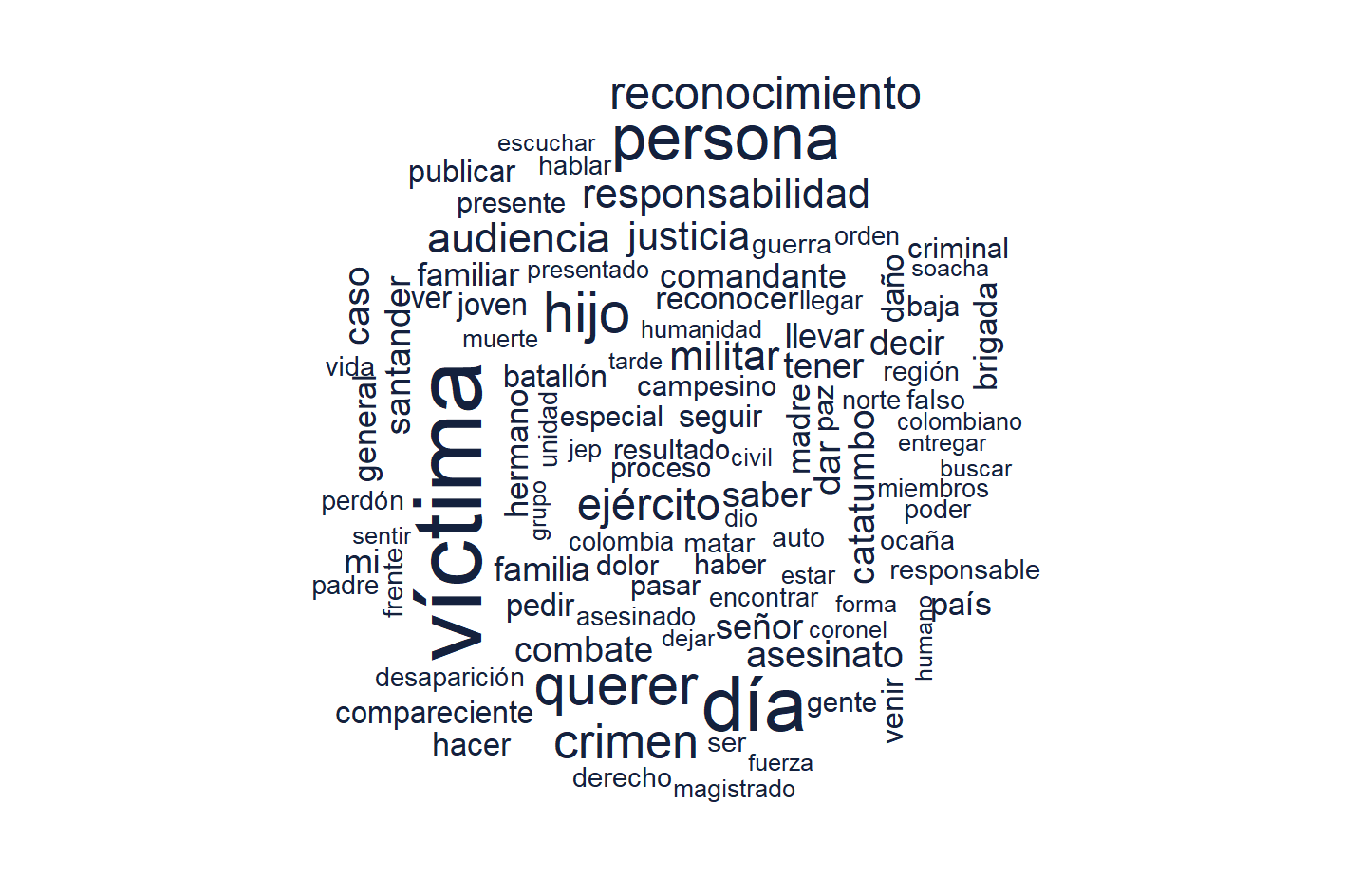}
        \caption{Norte de Santander: nube de Palabras.}
        \label{fig_nube_santander}
    \end{minipage}
\end{figure}

Al analizar las nubes de palabras generadas para los subcasos territoriales de Antioquia, Casanare, Costa Caribe, Huila, Meta y Norte de Santander (ver Figuras \ref{fig_nube_antioquia}, \ref{fig_nube_casanare}, \ref{fig_nube_costa_caribe}, \ref{fig_nube_huila}, \ref{fig_nube_meta} y \ref{fig_nube_santander}), emergen patrones lingüísticos diferenciados que reflejan matices en las formas en que se articulan el dolor, la verdad y la búsqueda de justicia en cada región. 


En el caso de Antioquia, a pesar de contar con un número limitado de videos, se aprecia con notable intensidad un léxico centrado en el reconocimiento del daño y la exigencia de verdad. Palabras como “víctima”, “justicia”, “familia”, “vida” y “militar” configuran un discurso centrado en la humanización de los afectados y en la responsabilidad estatal. La presencia de términos como “hijo”, “muchacho”, “niño”, “hermano” y “asesinada” aporta un tono profundamente personal y emocional, donde el sufrimiento se expresa en clave familiar e íntima. Esta configuración sugiere que, en Antioquia, el testimonio cumple una función reparadora, enfatizando el valor del reconocimiento como pilar de la justicia transicional.

En contraste, los subcasos de Casanare y Huila presentan un repertorio léxico más alineado con las dinámicas estructurales del conflicto armado. Predominan términos asociados a operaciones militares, jerarquías castrenses y hechos violentos, lo que refleja una narrativa menos centrada en las víctimas como sujetos individuales y más enfocada en el contexto institucional que permitió las violaciones. Este énfasis puede interpretarse como una reconstrucción del entorno bélico en el que se insertaron los crímenes, dando cuenta de las lógicas operativas que facilitaron su ejecución.

El subcaso de la Costa Caribe, por su parte, exhibe una configuración híbrida en la que se entrecruzan el lenguaje del reconocimiento y el de la reconciliación. Términos como “hermano”, “pueblo”, “hijo” y “responsabilidades” remiten tanto a la dimensión familiar del dolor como a una conciencia social del daño colectivo. La aparición destacada de la palabra “paz” sugiere que en esta región los discursos testimoniales están particularmente orientados hacia la reconstrucción del tejido social, resaltando el potencial restaurativo del proceso judicial.

En los subcasos de Meta y Norte de Santander, aunque el volumen de datos es reducido, se observan expresiones consistentes con los patrones previamente descritos: predominan los términos vinculados a la búsqueda de verdad, el reconocimiento del daño y la rendición de cuentas. Estas coincidencias refuerzan la idea de que, a pesar de las variaciones regionales, existe un núcleo común en las narrativas: la urgencia por esclarecer los hechos y avanzar en procesos de verdad y justicia que otorguen centralidad a las voces de las víctimas.

En conjunto, los subcasos analizados revelan tanto convergencias como singularidades en el modo en que los actores del Caso 03 de la JEP construyen sentido sobre su experiencia. Mientras algunos territorios enfatizan el carácter estructural del conflicto, otros recurren a un lenguaje más afectivo y relacional. Esta diversidad no solo enriquece la comprensión del fenómeno de los falsos positivos, sino que también ofrece claves contextuales fundamentales para el diseño de políticas de reparación diferenciadas y culturalmente sensibles.

Al aplicar las pruebas de hipótesis sobre los sentimientos expresados en los diferentes subcasos, se observa que Huila, Meta, Casanare y Norte de Santander reproducen un patrón similar al identificado en el análisis general. En estos territorios, las palabras con carga positiva son más frecuentes que las negativas, pero los sentimientos negativos, aunque menos utilizados, presentan una mayor intensidad emocional. Esta dualidad entre frecuencia y polaridad emocional sugiere que, si bien el lenguaje puede estar enmarcado en una narrativa de reconciliación, las emociones asociadas a la violencia, el dolor o la exigencia de justicia siguen siendo predominantes en términos de intensidad.

Sin embargo, Antioquia y la Costa Caribe presentan comportamientos diferenciados. En la Costa Caribe, los vectores de diferencia entre sentimientos positivos y negativos cumplen con el supuesto de normalidad según la prueba de Shapiro-Wilk. Por ello, se aplica la prueba \textit{t} de medias pareadas, la cual permite rechazar ambas hipótesis nulas con valores \textit{p} inferiores a 0.005. Estos resultados indican que tanto la frecuencia como la intensidad de los sentimientos negativos superan significativamente a los positivos, lo que posiciona a este subcaso como uno de los más cargados emocionalmente dentro del Caso 03, posiblemente reflejando una mayor tensión discursiva o una narrativa más marcada por el dolor y la denuncia.

En contraste, en Antioquia no se cumple el supuesto de normalidad en ninguno de los dos vectores. Por tanto, se recurre a la prueba de Wilcoxon para medianas pareadas, y en ambos casos se obtienen valores \textit{p} superiores a 0.1, lo que impide rechazar la hipótesis nula. Esto sugiere que, en promedio, los sentimientos negativos no predominan ni en número ni en intensidad respecto a los positivos. Así, Antioquia se configura como el subcaso con menor carga emocional negativa relativa, lo que puede estar relacionado con una narrativa más centrada en la verdad, la reparación o el reconocimiento institucional, en lugar de la denuncia directa.

Posteriormente, se replicó este análisis diferenciando entre víctimas y comparecientes en cada subcaso. Como se explicó anteriormente, debido al escaso número de videos categorizables en estas dos dimensiones, se excluyen del análisis los subcasos de Norte de Santander, Meta y los comparecientes de Antioquia, con el fin de preservar la validez estadística de las comparaciones. Al examinar los resultados, no se identifican diferencias significativas entre las narrativas de víctimas y comparecientes en términos de análisis de sentimientos, con una excepción importante: el subcaso de la Costa Caribe.

En este territorio, al segmentar el corpus de testimonios por rol, se observa un comportamiento diferencial. Para el grupo de víctimas, ambos vectores de diferencias cumplen con el supuesto de normalidad, por lo que se aplica nuevamente la prueba \textit{t} para la diferencia de medias. En este caso, el valor \textit{p} supera el umbral de significancia (mayor a 0.05), lo que impide rechazar la hipótesis nula. Esto implica que, entre las víctimas de la Costa Caribe, los sentimientos negativos no predominan ni en frecuencia ni en intensidad respecto a los positivos. Este hallazgo contrasta con los resultados generales del mismo subcaso y sugiere que los discursos de las víctimas, aunque emocionalmente complejos, pueden incorporar elementos de esperanza, dignificación o reconocimiento institucional que moderan la carga negativa expresada en sus testimonios. Esta divergencia entre el discurso de las víctimas y el de los comparecientes podría reflejar diferencias en las formas de narrar, en el posicionamiento ante la audiencia o en las expectativas depositadas en el proceso restaurativo impulsado por la JEP.

\subsection{Análisis de bigramas}

\begin{figure}[!htb]
    \centering
    \begin{minipage}{0.45\linewidth}
        \centering
        \includegraphics[width=\linewidth]{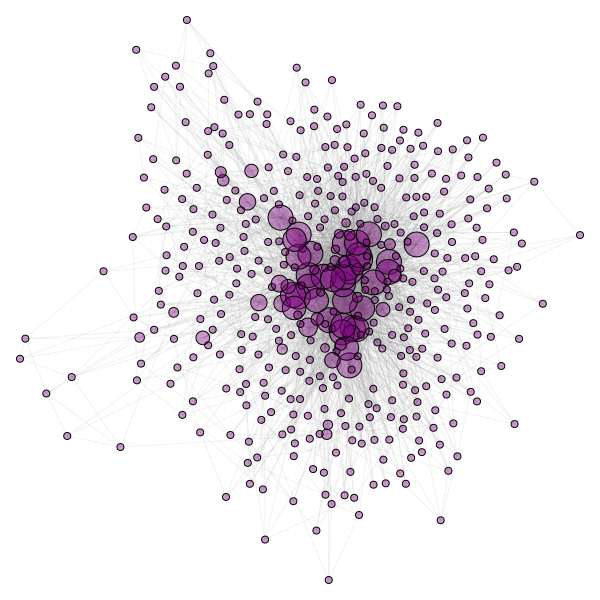} 
        \caption{Análisis general: grafo bigramas.}
        \label{fig:general-bigramas}
    \end{minipage}%
    \hfill
    \begin{minipage}{0.45\linewidth}
        \centering
        \includegraphics[width=\linewidth]{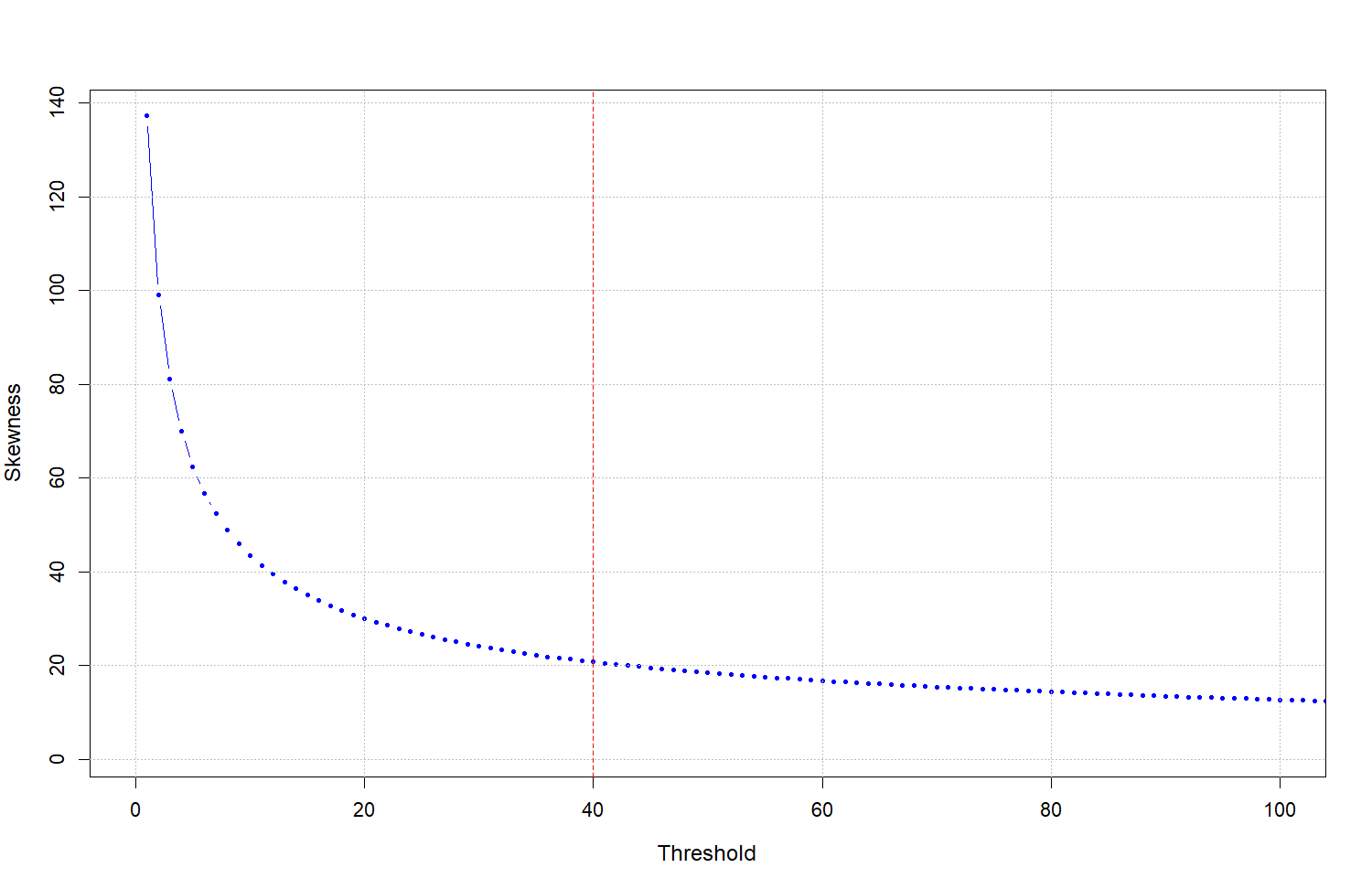} 
        \caption{Análisis general: límite bigramas.}
        \label{fig:threshold-bigramas}
    \end{minipage}
\end{figure}

Al generar los bigramas correspondientes al caso general, se decide conservar únicamente aquellos que aparecen más de 40 veces en el corpus, con el objetivo de enfocar el análisis en las asociaciones léxicas más recurrentes y significativas. Como se muestra en la Figura \ref{fig:threshold-bigramas}, al examinar la componente conexa principal de la red resultante, se evidencia que varios de los nodos presentan una centralidad de intermediación baja o nula, lo cual indica que estas palabras no desempeñan un papel relevante en la articulación de caminos entre otros términos del grafo.

Un patrón similar se observa al evaluar la centralidad propia, aunque en este caso la escasa intermediación resulta aún más pronunciada. Esta configuración sugiere que, si bien algunos términos aparecen con frecuencia, no necesariamente ocupan posiciones estratégicas dentro de la estructura semántica del discurso. Por esta razón, se opta por representar los nodos con un tamaño proporcional a su centralidad, lo que permite resaltar visualmente aquellas palabras que no solo son frecuentes, sino también estructuralmente relevantes en la red de significados. Esta estrategia facilita una interpretación más clara del papel que cumplen ciertos términos en la narrativa general, como se detalla en la Tabla \ref{tab:eigen}.

\begin{table}[!htb]
    \centering
    \begin{tabular}{lc}
        \hline
        Word & Eigen \\
        \hline
        batallón     & 1.000 \\
        comandante   & 0.827 \\
        sé           & 0.744 \\
        información  & 0.704 \\
        pronto       & 0.682 \\
        operaciones  & 0.680 \\
        brigada      & 0.674 \\
        recuerdo     & 0.670 \\
        \hline
    \end{tabular}
    \caption{Análisis general: palabras con mayor centralidad propia en la red de bigramas.}
    \label{tab:eigen}
\end{table}

Además del análisis visual de los bigramas, se calculan diversas métricas estructurales sobre la componente conexa principal de la red, cuyos resultados se presentan en la Tabla \ref{tab:measures}. La distancia media de 4.018 sugiere que, en promedio, cualquier par de términos relevantes está separado por más de cuatro pasos, lo que indica una red relativamente extensa y dispersa en términos semánticos. La media de grado (5.538) y su elevada desviación estándar (15.824) revelan una notable heterogeneidad en la conectividad: mientras algunos nodos actúan como términos altamente conectados —posiblemente estructuradores del discurso—, muchos otros presentan conexiones mínimas, lo que refuerza la idea de una red con núcleos léxicos específicos insertos en un mar de asociaciones más periféricas.

\begin{table}[!htb]
\centering
\begin{tabular}{lc}
\hline
Medida & Valor \\ \hline
Dist. media & 4.018 \\
Grado media & 5.538 \\
Grado desviación & 15.824 \\
Número clan & 20 \\
Densidad & 0.002 \\
Transitividad & 0.157 \\
Asortatividad & -0.101 \\ \hline
\end{tabular}
\caption{Análisis general: descripción de las medidas en la red de bigramas.}
\label{tab:measures}
\end{table}

La existencia de 20 clanes semánticos también apunta a una estructura fragmentada en subgrupos léxicos cohesionados, posiblemente reflejando distintos núcleos temáticos dentro del discurso general del Caso 03. Por su parte, la densidad de la red (0.002) es extremadamente baja, lo cual es consistente con redes lingüísticas grandes y sugiere que solo una fracción mínima de todas las posibles conexiones entre palabras está presente, reforzando la idea de dispersión temática.

La transitividad o coeficiente de agrupamiento (0.157) indica una baja pero no despreciable tendencia a la formación de triángulos, es decir, grupos de palabras que tienden a co-ocurrir entre sí, lo cual puede estar relacionado con estructuras narrativas repetitivas o frases comunes en los testimonios. Finalmente, la asortatividad negativa (-0.101) revela una ligera preferencia de los nodos con alto grado por conectarse con nodos de menor grado, lo que sugiere una estructura jerárquica donde algunos términos clave actúan como hubs que articulan muchas palabras menos conectadas, reforzando su centralidad discursiva dentro del análisis semántico. Estas métricas, en conjunto, confirman que el discurso analizado presenta una arquitectura semántica compleja, con vocabulario clave que organiza la narrativa alrededor de ejes temáticos bien definidos pero distribuidos de manera desigual.

El análisis de los 25 clústeres temáticos identificados mediante el algoritmo de detección de comunidades \textit{fast greedy} revela una estructura discursiva compleja y multifacética que refleja las principales dimensiones del Caso 03 de la JEP. Los temas identificados, presentados en la Tabla \ref{tab:temas_conflicto}, abarcan desde elementos operativos y jerárquicos del conflicto hasta aspectos emocionales, jurídicos y territoriales que atraviesan los testimonios analizados. Destacan especialmente tópicos como la manipulación de información y la ocurrencia de falsos positivos, la presencia de dudas o inconsistencias en los registros y declaraciones, así como la responsabilidad en crímenes de guerra, lo cual evidencia la centralidad de estos ejes en la narrativa del caso.

Asimismo, otros clústeres hacen referencia a la estructura de mando, los procedimientos operacionales y la logística militar, lo que reafirma el predominio de términos asociados a contextos castrenses ya observado en la red de bigramas y en el análisis de frecuencias. También emergen núcleos semánticos que expresan aspectos emocionales y de reconocimiento, lo que sugiere que, más allá de lo operativo y jurídico, el discurso recoge dimensiones humanas ligadas al dolor, la memoria y la búsqueda de verdad, en sintonía con las cargas afectivas observadas en el análisis de sentimiento. 

Esta diversidad temática, identificada dentro de una red léxica altamente dispersa, con baja densidad (0.002) y estructura modular bien definida (con 25 comunidades detectadas), confirma que el discurso judicial del Caso 03 se articula en torno a múltiples ejes narrativos. Estos ejes no solo reflejan la complejidad del conflicto armado colombiano, sino que también permiten rastrear patrones discursivos que conectan el accionar institucional, la experiencia de las víctimas y los mecanismos de justicia transicional, consolidando así una visión empírica más estructurada sobre la memoria judicial del caso.

\begin{table}[!htb]
    \centering
    \begin{tabular}{cl}
        \hline
        No. & Tema \\
        \hline
        1  & Manipulación de información y falsos positivos \\
        2  & Contexto social y territorial del conflicto \\
        3  & Inconsistencias en testimonios o declaraciones \\
        4  & Dudas o incertidumbre en registros \\
        5  & Operaciones y movimientos militares \\
        6  & Redes de informantes y extorsión \\
        7  & Estructura y operaciones de comando \\
        8  & Menores de edad y operaciones militares \\
        9  & Procesos y órdenes operacionales militares \\
        10 & Cadena de mando y operaciones militares \\
        11 & Grupos armados y presencia estatal \\
        12 & Acciones y confrontaciones armadas \\
        13 & Aspectos emocionales y de reconocimiento en el conflicto \\
        14 & Aspectos jurídicos y de derechos humanos \\
        15 & Operaciones y nombres de código en el conflicto \\
        16 & Geografía y terreno del conflicto \\
        17 & Procesos judiciales y beneficios penitenciarios \\
        18 & Logística y recursos económicos \\
        19 & Planificación y reflexión de actividades \\
        20 & Responsabilidad y reconocimiento en crímenes de guerra \\
        21 & Procedimientos y cronogramas judiciales \\
        22 & Geografía y eventos específicos del conflicto \\
        23 & Inteligencia y resultados de operaciones \\
        24 & Estrategias y crímenes de lesa humanidad \\
        25 & Jerarquía militar y representación legal \\
        \hline
    \end{tabular}
    \caption{Análisis general: tópicos.}
    \label{tab:temas_conflicto}
\end{table}

\begin{figure}[H]
    \centering
    \begin{minipage}{0.35\linewidth}
        \centering
        \includegraphics[width=\linewidth]{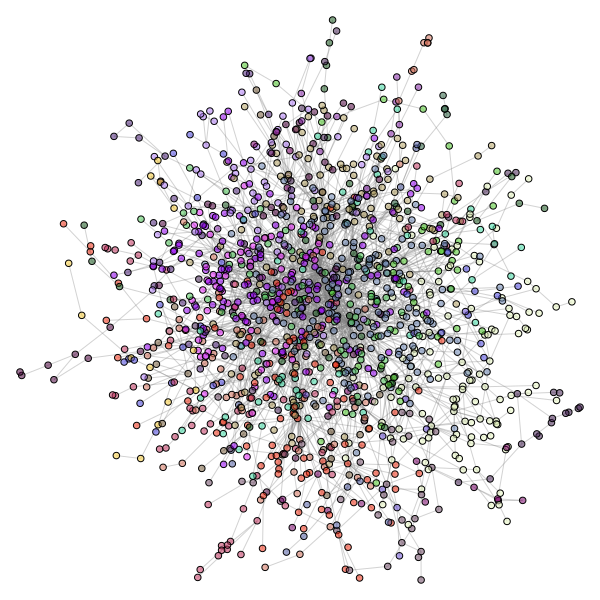} 
        \caption{Antioquia: grafo bigramas.}
        \label{fig_bi_antioquia}
    \end{minipage}%
    \hfill
    \begin{minipage}{0.35\linewidth}
        \centering
        \includegraphics[width=\linewidth]{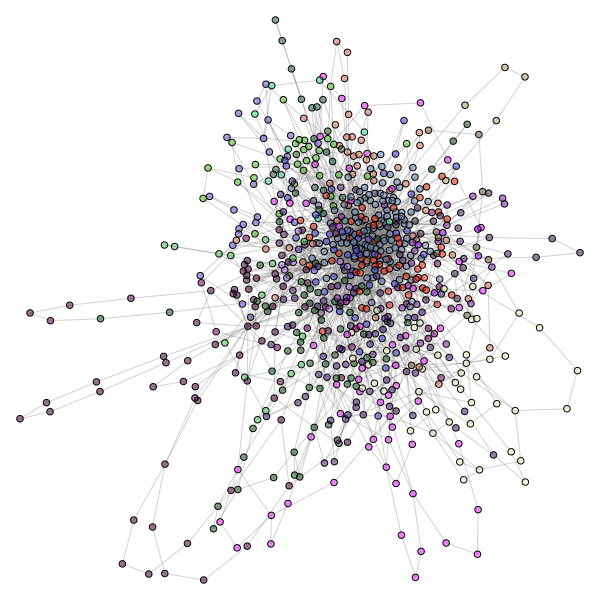}
        \caption{Casanare: grafo bigramas.}
        \label{fig_bi_casanare}
    \end{minipage}
\end{figure}

\begin{figure}[H]
    \begin{minipage}{0.4\linewidth}
        \centering
        \includegraphics[width=\linewidth]{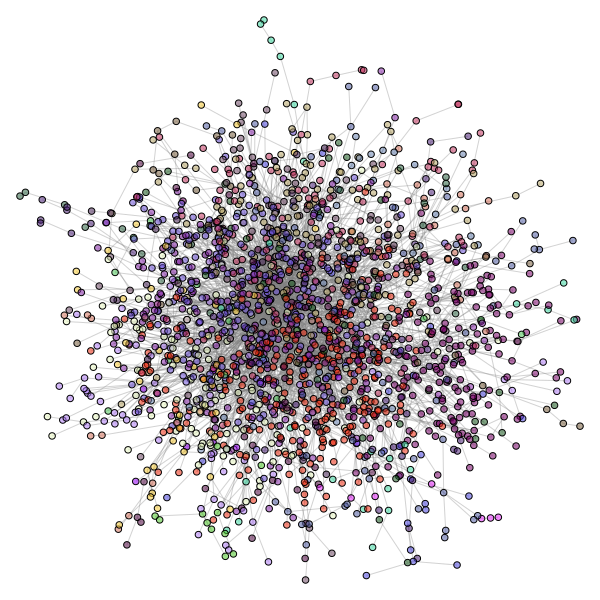}
        \caption{Costa Caribe: grafo bigramas.}
        \label{fig_bi_costa_caribe}
    \end{minipage}%
    \hfill
    \begin{minipage}{0.35\linewidth}
        \centering
        \includegraphics[width=\linewidth]{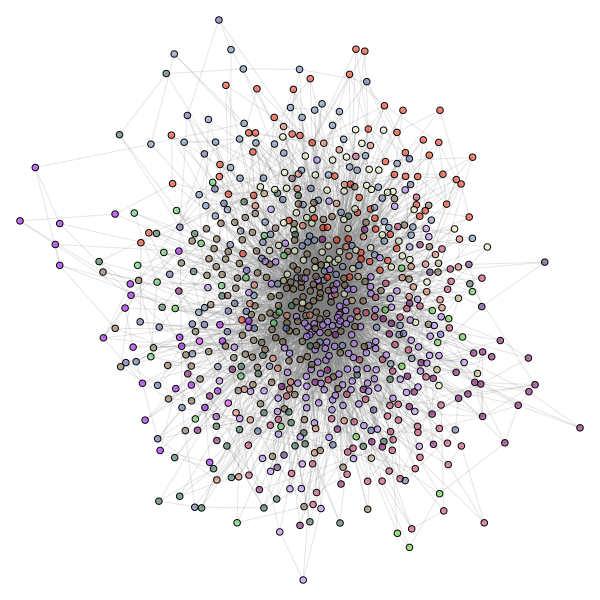}
        \caption{Huila: grafo bigramas.}
        \label{fig_bi_huila}
    \end{minipage}
\end{figure}

\begin{figure}[H]
    \begin{minipage}{0.35\linewidth}
        \centering
        \includegraphics[width=\linewidth]{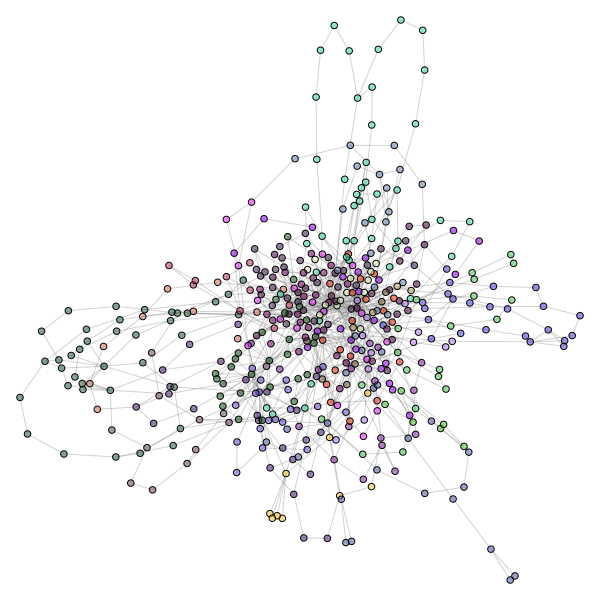}
        \caption{Meta: grafo bigramas.}
        \label{fig_bi_meta}
    \end{minipage}%
    \hfill
    \begin{minipage}{0.35\linewidth}
        \centering
        \includegraphics[width=\linewidth]{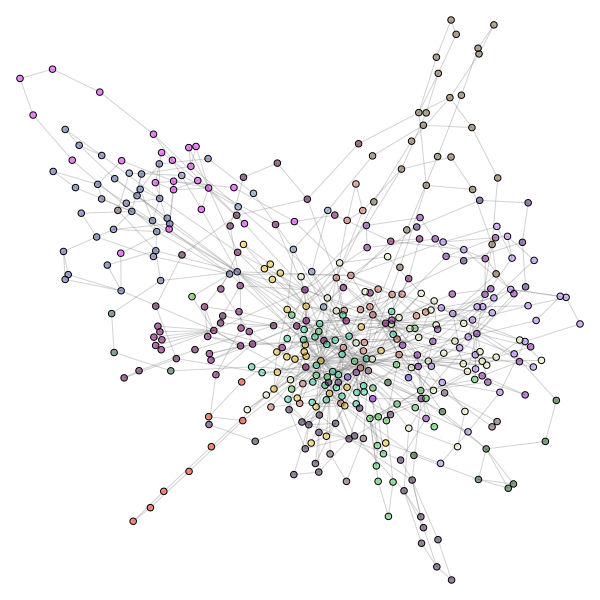}
        \caption{Norte de Santander: grafo bigramas.}
        \label{fig_bi_norte_santander}
    \end{minipage}
\end{figure}

A continuación, se realiza el análisis de las redes de bigramas para cada uno de los subcasos territoriales del Caso 03, con el objetivo de explorar las dinámicas semánticas y estructurales que emergen del discurso según el contexto regional. Las visualizaciones correspondientes se presentan en las Figuras \ref{fig_bi_antioquia}, \ref{fig_bi_casanare}, \ref{fig_bi_costa_caribe}, \ref{fig_bi_huila}, \ref{fig_bi_meta} y \ref{fig_bi_norte_santander}, donde se representa la red de coocurrencias para cada subcaso y se colorean los nodos según los clusters temáticos identificados mediante algoritmos de detección de comunidades.

En la Figura \ref{fig_bi_antioquia}, correspondiente al subcaso de Antioquia, se observa una red con baja densidad y estructura radial, con grupos temáticos poco conectados entre sí. Esta configuración sugiere una diversidad discursiva con menor cohesión entre tópicos, posiblemente asociada a una menor cantidad de testimonios y a la dispersión narrativa propia de los relatos individuales. Por el contrario, la Figura \ref{fig_bi_casanare}, que representa la red de Casanare, revela una estructura más densa y centralizada, con múltiples núcleos léxicos fuertemente conectados, lo cual sugiere un discurso más homogéneo y enfocado, posiblemente por la intensidad del caso en esta región y la abundancia de testimonios.

La red de la Costa Caribe, mostrada en la Figura \ref{fig_bi_costa_caribe}, se caracteriza por una distribución compacta de nodos y una segmentación clara entre clusters temáticos. La presencia de comunidades bien definidas indica un discurso orientado hacia tópicos específicos, como la verdad, el reconocimiento y la responsabilidad, coherente con las observaciones previas sobre la carga emocional y testimonial de esta región. En la Figura \ref{fig_bi_huila}, correspondiente a Huila, la red presenta una alta densidad de conexiones y una complejidad estructural notable, con numerosos nodos interrelacionados. Este patrón sugiere un discurso narrativo amplio y diversificado, en línea con el protagonismo del subcaso Huila dentro del Caso 03.

Por su parte, las redes correspondientes a Meta (Figura \ref{fig_bi_meta}) y Norte de Santander (Figura \ref{fig_bi_norte_santander}) presentan una configuración más dispersa y fragmentada, coherente con la limitada cantidad de videos disponibles para estos subcasos. Aunque las estructuras son más pequeñas, es posible identificar agrupamientos léxicos en torno a términos clave como “verdad”, “víctima” o “responsabilidad”, lo que muestra que, incluso en contextos con menor densidad discursiva, persisten patrones temáticos coherentes con los hallazgos generales.

En conjunto, el análisis de estas redes revela que los discursos construidos en torno a los subcasos territoriales no solo varían en volumen y complejidad, sino también en estructura semántica. Esta diversidad subraya la necesidad de abordar los procesos de justicia transicional desde una perspectiva diferenciada por territorio, reconociendo la forma en que el lenguaje configura distintas memorias del conflicto según el contexto regional. 


Al analizar la modularidad de las redes de bigramas construidas para cada subcaso territorial y compararlas con la del caso general, se observa una tendencia clara: los subcasos presentan niveles de modularidad considerablemente más altos. Mientras que la red general alcanza una modularidad de 0.469, los subcasos de Antioquia, Costa Caribe, Casanare, Meta y Norte de Santander exhiben valores superiores, con especial énfasis en Meta (0.740) y Norte de Santander (0.756). Esta diferencia sugiere que, al desagregar el corpus según los contextos regionales, emergen estructuras semánticas más coherentes y compactas, lo cual refleja una mayor cohesión temática en cada subgrupo. En otras palabras, la partición por subcasos no solo se justifica por criterios analíticos o administrativos, sino que también encuentra respaldo en la organización interna del lenguaje: los discursos regionales revelan patrones más consistentes y bien definidos que se diluyen en el análisis agregado.

La modularidad, como medida de calidad en la detección de comunidades léxicas, refuerza así la pertinencia de este enfoque segmentado. En particular, los subcasos de Antioquia y Costa Caribe sobresalen no solo por su alta modularidad (0.695 y 0.650 respectivamente), sino también por la claridad temática observada en sus redes. Esta consistencia estructural coincide con los hallazgos del análisis de sentimiento y de tópicos realizados previamente, donde ambos subcasos mostraban narrativas distintivas en cuanto a carga emocional y foco discursivo. Por lo tanto, la comparación de modularidad no solo valida la decisión metodológica de trabajar con subcasos, sino que también aporta evidencia empírica sobre la heterogeneidad del conflicto armado colombiano y sobre la necesidad de aproximaciones diferenciadas para comprender sus múltiples dimensiones narrativas y afectivas.

El análisis temático del subcaso de Antioquia revela una red discursiva particularmente estructurada y coherente, lo cual coincide con los altos niveles de modularidad observados previamente. A partir de la clasificación automática de los bigramas, se identifican 25 clústeres semánticos que abordan de manera articulada múltiples dimensiones del conflicto armado en esta región (ver Tabla \ref{tab:temas_conflicto_antioquia}). Destaca la centralidad de las víctimas y sus familiares, no solo como sujetos de daño, sino también como protagonistas activos en los procesos de verdad, justicia y reparación. Esta narrativa se entrelaza con temas sobre la responsabilidad de las fuerzas armadas, los crímenes de lesa humanidad y los patrones de macrocriminalidad asociados a los llamados “falsos positivos”.
    
\begin{table}[!htb]
\centering
\begin{tabular}{cl}
\hline
No. & Tópico \\
\hline
1 &Víctimas y representación en la JEP\\
2 &Crímenes de lesa humanidad y responsabilidad\\
3 &Conflicto armado y dinámica de violencia\\
4 &Patrones de macrocriminalidad y sistemática\\
5 &Procedimientos y participación en audiencias\\
6 &Nombres propios y personalidades involucradas\\
7 &Esperanza y búsqueda de la verdad\\
8 &Impacto social y comunitario\\
9 &Procesos de paz y justicia restaurativa\\
10 &Petición de perdón y reconocimiento de responsabilidad\\
11 &Agradecimientos y cierre de intervenciones\\
12 &Nombres propios y personalidades secundarias\\
13 &Rol del ejército y agentes estatales en el conflicto\\
14 &Solidaridad y reconocimientos institucionales\\
15 &Vulnerabilidad de las víctimas y afectados\\
16 &Falsos positivos y asesinatos extrajudiciales\\
17 &Proceso legal y marco jurídico\\
18 &Búsqueda de justicia y reparación\\
19 &Impacto personal y familiar de la violencia\\
20 &Reflexiones y mensajes finales\\
21 &Presencia y reconocimiento de asistentes\\
22 &Impacto en la salud y comunidad\\
23 &Menores de edad y reclutamiento\\
24 &Dolor y daño causado por el conflicto\\
25 &Estructura militar y operaciones en Antioquia\\
\hline
\end{tabular}
\caption{Antioquia: tópicos.}
\label{tab:temas_conflicto_antioquia}
\end{table}

Los tópicos detectados también capturan elementos institucionales y procedimentales, tales como la representación ante la JEP, las dinámicas propias de las audiencias, y los procesos judiciales en curso. Se identifican nombres propios, unidades militares, y agentes estatales, lo cual da cuenta de un discurso orientado a individualizar responsabilidades y exigir rendición de cuentas. Asimismo, el análisis revela componentes profundamente humanos y emocionales: el dolor causado por la violencia, la búsqueda de verdad, las peticiones de perdón, y los mensajes de cierre cargados de esperanza o agradecimientos, los cuales refuerzan el carácter restaurativo de estas intervenciones.

Finalmente, se destacan temas que abordan la vulnerabilidad estructural de las víctimas, la afectación comunitaria, el reclutamiento de menores y las secuelas físicas y emocionales del conflicto. Todo ello refleja una narrativa que articula demandas de justicia con llamados a la reconciliación, enmarcados dentro de un discurso institucional que reconoce tanto los crímenes como los procesos de reparación. En conjunto, los resultados del subcaso de Antioquia confirman la utilidad del análisis computacional para descomponer con precisión los ejes narrativos presentes en los testimonios, y evidencian cómo, en esta región, el reconocimiento de las víctimas y la memoria del daño ocupan un lugar central en la construcción de verdad.

El análisis temático del subcaso de la Costa Caribe revela una narrativa densa, compleja y fuertemente estructurada, reflejo de una región particularmente impactada por las dinámicas del conflicto armado y los crímenes de Estado (ver Tabla \ref{tab:temas_conflicto_costa_caribe}). Los 25 clústeres obtenidos en la red semántica de bigramas capturan con precisión tanto las dimensiones estructurales del fenómeno de los falsos positivos como sus consecuencias personales, familiares y comunitarias. Uno de los aspectos más notables de este subcaso es la centralidad que adquieren las comunidades indígenas y autoridades étnicas, que no solo aparecen como víctimas directas, sino también como colectivos históricamente vulnerados, cuyas cosmovisiones, territorios y liderazgos han sido profundamente afectados.

\begin{table}[!htb]
\centering
\begin{tabular}{cl}
\hline
No. & Tópico \\
\hline
1 &Historia y contexto de las ejecuciones extrajudiciales en Colombia\\
2 &Patrones y estructura de prácticas criminales\\
3 &Impacto social y legal de las ejecuciones extrajudiciales\\
4 &Geografía del conflicto y actores regionales\\
5 &Impacto personal y familiar de las víctimas\\
6 &Comunidades indígenas y su relación con el conflicto\\
7 &Agentes del Estado involucrados en ejecuciones extrajudiciales\\
8 &Vulnerabilidad y condiciones de las víctimas\\
9 &Procedimientos y medidas judiciales\\
10 &Metodología y fases de investigación judicial\\
11 &Reacciones emocionales y personales ante las ejecuciones\\
12 &Proceso judicial y reconocimiento de las víctimas\\
13 &Consecuencias personales y trauma de las ejecuciones\\
14 &Involucramiento del Ejército y estructuras militares\\
15 &Operaciones militares y alias de actores armados en municipios específicos\\
16 &Reparación y conocimiento del daño a las víctimas\\
17 &Condiciones y efectos de la violencia en las comunidades\\
18 &Responsabilidad y admisión de actos criminales\\
19 &Territorios ancestrales y espacios críticos del conflicto\\
20 &Impacto y necesidades de las familias afectadas\\
21 &Reconocimiento de responsabilidades y falsos positivos\\
22 &Operativos y señalamientos falsos\\
23 &Operaciones militares y falsos combates\\
24 &Marco legal y procesos de paz y justicia\\
25 &Presentación de muertes y asesinatos en combate\\
\hline
\end{tabular}
\caption{Antioquia: tópicos.}
\label{tab:temas_conflicto_costa_caribe}
\end{table}

Los tópicos identifican de manera sistemática la participación de agentes estatales, especialmente miembros del Ejército, en prácticas de encubrimiento, operaciones falsas y ejecuciones extrajudiciales, enmarcadas en lógicas perversas de acumulación de resultados. La alusión a alias, operativos específicos y lugares concretos de la geografía regional refuerza la idea de un conflicto situado territorialmente, donde la violencia no fue aleatoria, sino estructurada. Temas como los procedimientos judiciales, la reparación y el reconocimiento de responsabilidades emergen como ejes discursivos clave, lo que indica una narrativa que no se limita a la denuncia, sino que también articula demandas concretas de justicia y reparación integral.

En términos emocionales, los discursos de este subcaso se distinguen por un tono de fuerte carga afectiva, donde el trauma, la indignación y la búsqueda de verdad se entrelazan. El análisis cuantitativo corrobora este patrón: la intensidad de los sentimientos negativos supera significativamente la de los positivos, y los resultados de las pruebas estadísticas indican que este es uno de los pocos subcasos donde los sentimientos negativos no solo son más intensos, sino también más frecuentes. Este hallazgo, junto con la elevada modularidad de la red semántica (0.650), sugiere una narrativa fuertemente cohesionada alrededor de la denuncia, el dolor, la exigencia de justicia y el reclamo de verdad histórica. En conjunto, los hallazgos posicionan al subcaso de la Costa Caribe como uno de los más estructurados discursivamente y con mayor carga emocional negativa dentro del caso 03 de la JEP, lo que refuerza la relevancia de su análisis específico en el marco de la justicia transicional.

El análisis de modularidad en las redes de bigramas segmentadas por víctimas y comparecientes arroja resultados reveladores sobre la estructura del discurso en el caso 03. Como se observa en la Tabla \ref{tab:bigramas_regiones_completo}, al comparar la modularidad del discurso general con la obtenida al separar las narrativas según el rol de los participantes, se identifican comportamientos divergentes entre los distintos subcasos. En algunos de ellos, como Costa Caribe y Casanare, la segmentación conduce a un aumento considerable en la modularidad —por ejemplo, en Costa Caribe se eleva de 0.650 a 0.769 para las víctimas y 0.716 para los comparecientes— lo que indica una mayor cohesión temática cuando se analiza cada grupo por separado. Esta fragmentación temática sugiere que las experiencias, las formas de narrar los hechos y los énfasis discursivos difieren sustancialmente entre víctimas y comparecientes en esas regiones, posiblemente debido a las dinámicas particulares del conflicto y las distintas posiciones subjetivas que ocupan en el relato.

Sin embargo, este patrón no es uniforme. En el caso de Huila, por ejemplo, la modularidad mejora levemente al analizar por separado las narrativas (de 0.445 a 0.582 para víctimas), pero en el caso de los comparecientes apenas cambia (0.436), lo que indica una menor segmentación temática o una mayor dispersión discursiva en ese grupo. A nivel general, el caso agregado presenta una modularidad de 0.469, mientras que al separar por roles se observa una mejora moderada (0.596 para víctimas y 0.422 para comparecientes). No obstante, en Antioquia la narrativa de los comparecientes no se analiza debido al bajo número de observaciones, aunque la modularidad de las víctimas es ligeramente superior a la del total (0.714 frente a 0.695), sugiriendo una narrativa más estructurada dentro de ese grupo.

En conjunto, estos hallazgos indican que, si bien en algunos contextos regionales resulta metodológicamente pertinente desagregar las narrativas, en otros, el discurso compartido entre víctimas y comparecientes refleja una mayor interdependencia temática. Este comportamiento puede interpretarse como evidencia de una estructura narrativa común en torno a los hechos, lo que podría estar asociado a los formatos institucionales de las audiencias, las dinámicas procesales de la JEP o incluso a la forma en que se ha construido colectivamente la memoria del conflicto. Así, la modularidad no solo actúa como indicador de cohesión semántica, sino también como una herramienta para explorar las distancias y proximidades discursivas entre actores enfrentados por la guerra.    

\begin{table}[!htb]
\centering
\begin{tabular}{lccc}
\hline
Subcaso & Todos & Comparecientes & Víctimas \\ \hline
Antioquia          & 0.695 & -    & 0.714 \\ 
Casanare           & 0.573 &0.507 & 0.689 \\ 
Costa Caribe       & 0.650 & 0.716 & 0.769 \\ 
Huila              & 0.445 & 0.436 & 0.582 \\ 
General            & 0.469 & 0.422 & 0.596 \\ \hline
\end{tabular}
\caption{Comparación de modularidad en redes de bigramas según subcaso y rol de los participantes (víctimas y comparecientes).}
\label{tab:bigramas_regiones_completo}
\end{table}

\subsection{Análisis de \textit{skipgramas}}

\begin{figure}[!htb]
    \begin{minipage}{0.45\linewidth}
        \centering
        \includegraphics[width=\linewidth]{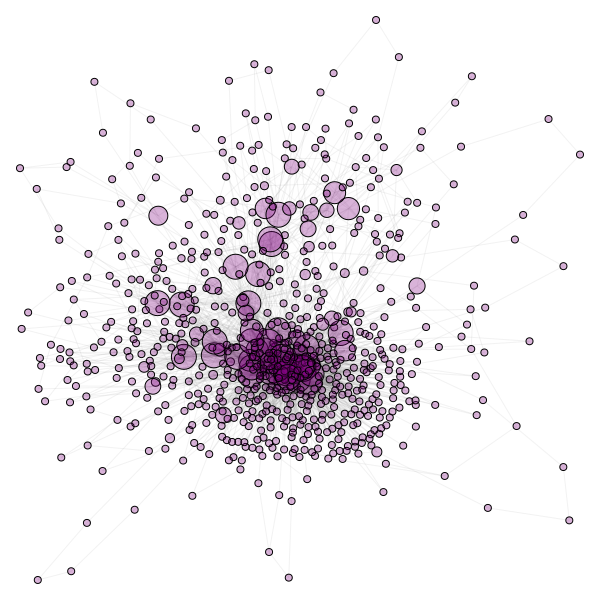}
        \caption{Análisis general: grafo \textit{skipgramas}.}
        \label{fig:general-skipgramas}
    \end{minipage}%
    \hfill
    \begin{minipage}{0.45\linewidth}
        \centering
        \includegraphics[width=\linewidth]{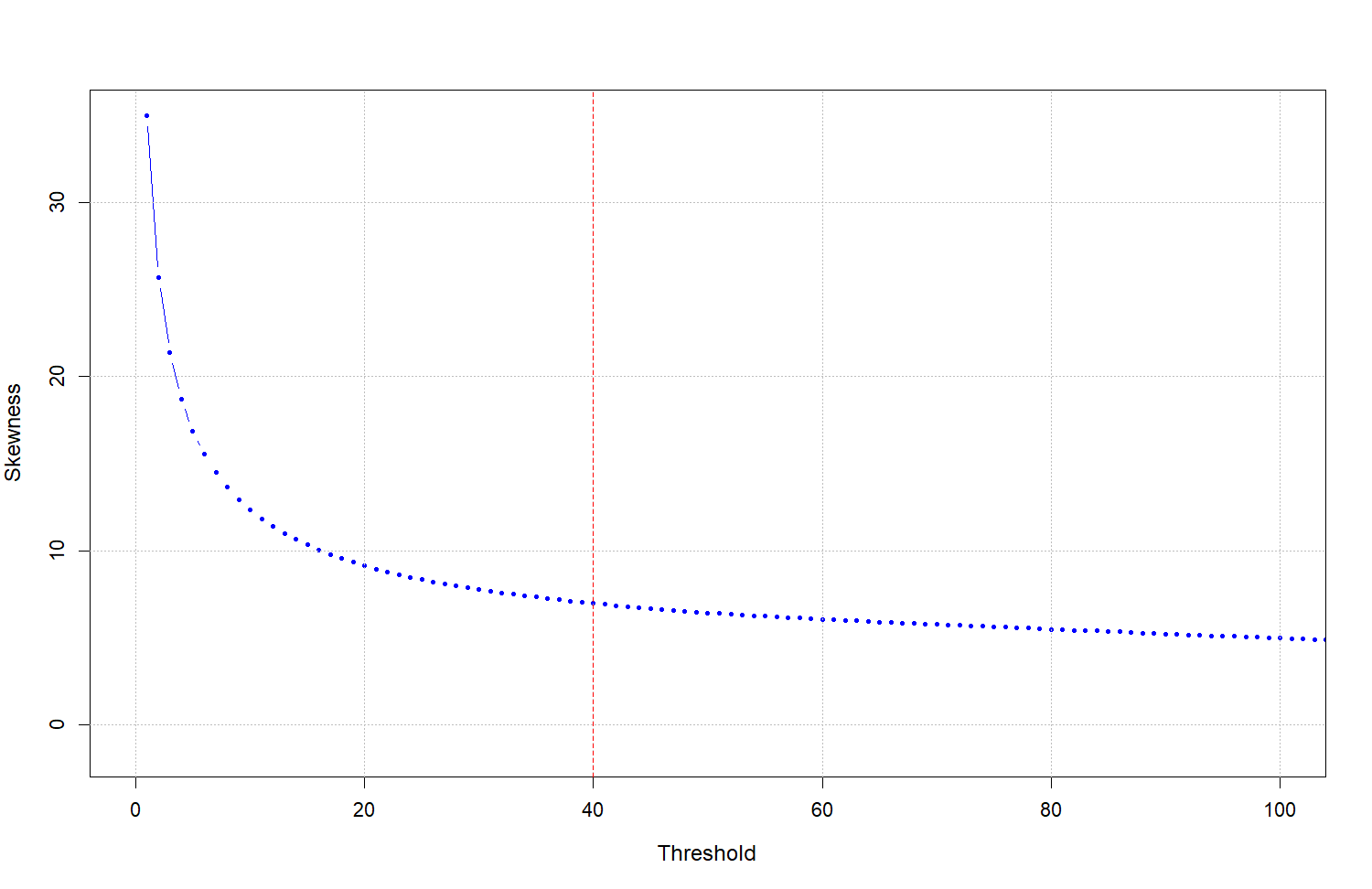}
        \caption{Anáisis general: límite \textit{skipgramas}.}
        \label{fig:threshold-skipgramas}
    \end{minipage}
\end{figure}

Al generar la red de \textit{skipgramas} del caso general, se opta, al igual que en el análisis de bigramas, por conservar únicamente aquellas combinaciones de palabras que aparecen más de 40 veces en el corpus. Esta decisión se justifica visualmente en la Figura \ref{fig:threshold-skipgramas}, donde se observa que a partir del umbral de 40 la asimetría de la distribución de frecuencias (sesgo) comienza a estabilizarse, lo que indica una reducción significativa del ruido léxico. La red resultante se presenta en la Figura \ref{fig:general-skipgramas}, donde los nodos están escalados según su centralidad de intermediación.

A diferencia de los bigramas, la red de skipgramas muestra una estructura más concentrada en torno a un núcleo denso con múltiples nodos altamente conectados. Esto sugiere que ciertos términos desempeñan un papel clave como intermediarios semánticos entre distintas partes del discurso, contribuyendo a la cohesión temática. Además, el mayor número de nodos con centralidad alta resalta la riqueza semántica de esta representación, ya que permite capturar conexiones más sutiles entre palabras no contiguas que emergen del testimonio oral.

En este sentido, los skipgramas revelan una estructura discursiva más interconectada y posiblemente más útil para captar relaciones latentes en el lenguaje de las audiencias, en comparación con los bigramas, que tienden a reflejar relaciones más literales y locales. Esta diferencia estructural entre ambas representaciones tiene implicaciones metodológicas importantes para la extracción de tópicos y la caracterización narrativa del caso 03.

El análisis de los tópicos extraídos mediante la red de skipgramas permite identificar un conjunto diverso de temas que reflejan la complejidad del caso 03 desde múltiples dimensiones (ver Tabla \ref{tab:clusters}). Predominan tópicos relacionados con la estructura y operatividad militar, como se observa en los clústeres 1, 4, 5, 14, 16 y 19, que abordan desde la organización y funciones de batallones y Gaulas, hasta la logística operativa y los movimientos jerárquicos dentro de las unidades. Estas agrupaciones evidencian un discurso fuertemente anclado en la dimensión institucional y táctica del conflicto armado.

Simultáneamente, se destacan aspectos jurídicos y de responsabilidad institucional, como lo indican los clústeres 9, 10, 12, 13, 17 y 18, los cuales tratan temas vinculados con la participación de actores armados, la admisión de responsabilidades, el desarrollo de audiencias, y los marcos legales aplicados por la Jurisdicción Especial para la Paz. Estas agrupaciones aportan una mirada crítica sobre el proceso judicial y las herramientas de justicia transicional implementadas en el caso.

\begin{table}[!htb]
    \centering
    \begin{tabular}{cl}
        \hline
        No. & Tópico \\
        \hline
        1  & Estructura y operaciones militares \\
        2  & Información personal y roles de los implicados \\
        3  & Testimonios y recuerdos de las operaciones \\
        4  & Procedimientos y protocolos militares \\
        5  & Dinámicas y movimientos de comandantes y unidades \\
        6  & Consecuencias y sanciones disciplinarias \\
        7  & Personajes específicos en el caso \\
        8  & Zonas geográficas relevantes en el conflicto \\
        9  & Responsabilidad y participación en el conflicto \\
        10 & Procedimientos judiciales y reconocimientos de víctimas \\
        11 & Procesos de reconciliación y reparación territorial \\
        12 & Funciones e intervención judicial \\
        13 & Procesos judiciales y formación militar \\
        14 & Operaciones y estructuras de batallones \\
        15 & Iniciativas de solidaridad y política \\
        16 & Organización y dirección de Gaulas \\
        17 & Marco legal y procedimientos especiales \\
        18 & Procesos de justicia equitativa \\
        19 & Equipamiento y logística militar \\
        \hline
    \end{tabular}
    \caption{Análisis general: tópicos.}
    \label{tab:clusters}
\end{table}

No obstante, emergen también clústeres centrados en la dimensión humana y restaurativa del proceso, particularmente los clústeres 2, 3, 7, 11 y 15. Estos tópicos contienen referencias a testimonios personales, recuerdos, actos de reconciliación, reparación territorial, y expresiones de solidaridad y reconocimiento entre actores, lo que sugiere un componente afectivo y ético dentro del relato. En este sentido, la red de skipgramas no solo revela estructuras de poder y responsabilidad, sino también vínculos emocionales y sociales que atraviesan la experiencia del conflicto.

En conjunto, la clasificación temática obtenida permite analizar el Caso 03 desde una perspectiva integral que articula lo operativo, lo legal y lo social. Esta multiplicidad de enfoques refuerza el valor de los skipgramas como herramienta para capturar las relaciones latentes entre los actores, discursos e instituciones implicadas en la justicia transicional colombiana.

Al analizar los valores de modularidad obtenidos para los diferentes subcasos en la red de skipgramas, se observa una tendencia consistente con lo reportado en el análisis de bigramas: los discursos tienden a presentar una mayor cohesión interna cuando se segmentan por regiones. No obstante, en el caso de los skipgramas, la magnitud de las diferencias entre el caso general y los subcasos es aún más marcada, lo cual refuerza la utilidad de este enfoque para captar relaciones semánticas no contiguas, revelando estructuras discursivas más complejas y específicas.

La modularidad global de la red de skipgramas en el caso general es de 0.360, un valor considerablemente inferior al de la mayoría de los subcasos, lo que indica una menor cohesión temática cuando se agrupan todos los discursos en un solo conjunto. En contraste, subcasos como Meta (0.687), Antioquia (0.616) y la Costa Caribe (0.596) muestran modularidades elevadas, lo que sugiere una organización interna más clara y agrupamientos semánticos bien definidos al interior de cada región. Este patrón también se observa en Norte de Santander (0.577) y Casanare (0.489), aunque con una menor intensidad. Huila, por su parte, presenta un valor de modularidad relativamente bajo (0.337), muy cercano al del caso general, lo que podría interpretarse como una menor diferenciación temática en los discursos de esa región.

Comparado con los valores reportados en la red de bigramas, donde la modularidad general fue de 0.469 y los subcasos oscilaron entre 0.445 (Huila) y 0.756 (Norte de Santander), se aprecia que, si bien los bigramas capturan bien la cohesión temática, los skipgramas ofrecen una ventaja adicional al capturar relaciones léxicas más flexibles que no dependen de la contigüidad directa. Esta flexibilidad permite una representación más matizada de las co-ocurrencias discursivas, lo que se traduce en agrupamientos más informativos en algunos subcasos, como en Meta y la Costa Caribe.

En suma, el análisis de modularidad en las redes de skipgramas confirma que el discurso asociado al Caso 03 de la JEP adquiere una estructura más homogénea y significativa cuando se examina por subcasos territoriales. Este hallazgo no solo valida la pertinencia de la segmentación regional para el análisis lingüístico, sino que también evidencia cómo distintas regiones construyen y articulan sus narrativas en torno al conflicto armado con distintos niveles de cohesión interna, lo que puede reflejar diferencias en las dinámicas del conflicto, la naturaleza de los testimonios o el grado de avance del proceso restaurativo en cada zona.

La clasificación temática derivada del análisis de skipgramas para el subcaso de Antioquia (Tabla \ref{tab:temas_conflicto_antioquia_skip}) evidencia una narrativa profundamente centrada en las experiencias de las víctimas y el impacto social del conflicto armado. Los tópicos identificados incluyen temas cruciales como las ejecuciones extrajudiciales, las desapariciones forzadas, la búsqueda de justicia y verdad, y los procesos de restauración y reparación social. En particular, destacan aquellos que abordan el sufrimiento de las comunidades, la descomposición del tejido familiar y el rol fundamental de la JEP como escenario de reconocimiento y memoria.

\begin{table}[!htb]
    \centering
    \begin{tabular}{cl}
        \hline
        No. & Tópico \\
        \hline
        1  & Impacto de las ejecuciones extrajudiciales en la vida familiar y comunitaria \\
        2  & Ejecuciones extrajudiciales y desapariciones forzadas \\
        3  & Identificación de víctimas y personas involucradas \\
        4  & Derechos humanos y violaciones en el contexto del conflicto \\
        5  & Restauración y reparación social \\
        6  & Sistema de justicia transicional \\
        7  & Aspectos militares y recuperación de cuerpos \\
        8  & Procesos judiciales y recolección de testimonios \\
        9  & Detalles procedimentales en audiencias \\
        10 & Carácter exhaustivo de la investigación \\
        11 & Elementos simbólicos o incidentales \\
        12 & Conflicto armado y violencia interna \\
        13 & Figuras legales y administrativas asociadas a la JEP \\
        14 & Instituciones militares y responsabilidad en crímenes \\
        15 & Grupos armados y dinámicas del conflicto \\
        16 & Relatos de soldados y asesinatos \\
        17 & Audiencias públicas y testimonios de las víctimas \\
        18 & Educación y entornos juveniles \\
        19 & Víctimas y búsqueda de justicia y verdad \\
        20 & Geografía del conflicto y casos específicos \\
        21 & Solidaridad y emociones en el proceso de justicia \\
        22 & Impacto del conflicto en municipios específicos \\
        \hline
    \end{tabular}
\caption{Antioquia: tópicos.}
\label{tab:temas_conflicto_antioquia_skip}
\end{table}

Se abordan también dimensiones jurídicas y procedimentales, como la responsabilidad institucional de las Fuerzas Militares, los detalles técnicos de las audiencias, la recopilación de testimonios y los aspectos legales vinculados a la justicia transicional. Además, se incorporan aspectos simbólicos y emocionales del proceso judicial, incluyendo expresiones de solidaridad, participación juvenil y referencias geográficas específicas, que sitúan el conflicto en un marco territorial concreto.

En comparación con el análisis de bigramas, el enfoque de skipgramas permitió una mayor granularidad en la identificación de tópicos, lo cual se refleja también en el valor de modularidad reportado para esta red (0.616). Aunque este valor es ligeramente menor que el de la red de bigramas en el mismo subcaso (0.695), los tópicos derivados del modelo de skipgramas revelan una narrativa más matizada, donde las relaciones semánticas no contiguas entre términos permiten capturar temas más densos y variados. Esta ligera disminución en la modularidad puede interpretarse no como una pérdida de coherencia, sino como una ganancia en complejidad discursiva: se integran múltiples capas del conflicto —legales, militares, familiares y emocionales— en un mismo espacio semántico.

En el caso de la Costa Caribe, como se observa en la Tabla \ref{tab:temas_conflicto_costa_caribe_skip}, el análisis temático a partir de la red de skipgramas en este subcaso revela una narrativa profundamente marcada por las consecuencias del conflicto armado en el tejido social y comunitario. Entre los temas más destacados se encuentran el impacto social y cultural de las ejecuciones extrajudiciales, el trauma emocional en las víctimas y sus familias, y las desapariciones forzadas, así como la búsqueda activa de justicia y verdad.

Esta región presenta una fuerte presencia de tópicos relacionados con los procesos judiciales, tanto en términos de participación en las audiencias de la Jurisdicción Especial para la Paz (JEP) como en el reconocimiento institucional de las víctimas. Se hace énfasis en los procedimientos legales, las violaciones al derecho internacional humanitario y los mecanismos de justicia transicional, todos ellos vinculados a un contexto de macrocriminalidad y responsabilidad penal por crímenes de guerra y de lesa humanidad.

Asimismo, se identifican temas orientados al papel del activismo comunitario y la defensa de derechos. Conceptos como liderazgo social, resistencia comunitaria y solidaridad emergen con fuerza en el discurso, mostrando cómo las comunidades no solo fueron víctimas del conflicto, sino también agentes activos en los procesos de memoria, denuncia y reconstrucción.

En contraste con lo observado en el análisis de bigramas para esta misma región —donde predominaban expresiones operativas y de responsabilidad militar—, los skipgramas permiten acceder a relaciones semánticas más complejas y distantes, lo que enriquece la detección de tópicos vinculados al dolor social, la movilización colectiva y el reconocimiento institucional. Esta mayor diversidad discursiva se ve reflejada en el valor de modularidad obtenido (0.596), apenas ligeramente inferior al del análisis con bigramas (0.650), lo que sugiere que ambas metodologías capturan estructuras cohesivas pero con matices diferentes: mientras los bigramas tienden a reflejar lo explícito y repetido, los skipgramas revelan conexiones más profundas, dispersas y a menudo más significativas a nivel temático.

En conjunto, el discurso de la Costa Caribe aparece como uno de los más articulados en términos de denuncia, reparación y resistencia. El análisis de skipgramas refuerza la centralidad de las víctimas como sujetos políticos y sociales activos, al tiempo que permite observar cómo se articulan demandas de justicia con memorias del dolor y apuestas por la transformación social.

\begin{table}[!htb]
\centering
\begin{tabular}{cl}
\hline
No. & Tópico \\
\hline
1& Impacto social y cultural de las ejecuciones extrajudiciales\\
2& Impacto personal y emocional en las víctimas y sus familias\\
3& Procedimientos y participantes en las audiencias de la JEP\\
4& Efectos de las acciones militares en comunidades locales\\
5& Procesos y mecanismos de justicia transicional\\
6& Aspectos jurídicos y sociales relacionados con el conflicto\\
7& Derechos humanos y crímenes cometidos en el marco del conflicto\\
8& Consecuencias y búsqueda de justicia por parte de las víctimas\\
9& Denuncias y reconocimiento de responsabilidades en el conflicto\\
10& Crímenes de guerra y de lesa humanidad\\
11& Conceptos de macrocriminalidad y crímenes extendidos\\
12& Identificación de víctimas y perpetradores\\
13& Aspectos sociales y legales relacionados con las muertes\\
14& Derecho internacional humanitario y justicia transicional\\
15& Rol de las fuerzas armadas y paramilitares en el conflicto\\
16& Dinámicas de los grupos armados y su impacto\\
17& Prácticas de exterminio y su impacto en la confianza social\\
18& Impacto del conflicto en los pueblos indígenas\\
19& Impacto del conflicto en las comunidades locales\\
20& Violencia y persecución en el marco del conflicto armado\\
21& Procedimientos judiciales y solidaridad social\\
22& Cambios y consecuencias de las acciones ilegales\\
23& Desapariciones forzadas y la búsqueda de los desaparecidos\\
24& Liderazgo comunitario y resistencia social\\
\hline
\end{tabular}
\caption{Costa Caribe: tópicos.}
\label{tab:temas_conflicto_costa_caribe_skip}
\end{table}

Como se observa en la Tabla \ref{tab:skipgramas_regiones_completo}, el análisis de redes de skipgramas desagregado por rol (víctimas y comparecientes) muestra una leve mejora en los valores de modularidad para los subcasos de Antioquia y la Costa Caribe en comparación con los resultados agregados. En el caso de Antioquia, el valor de modularidad se mantiene constante en 0.616, dado que no se cuenta con datos suficientes para construir una red exclusivamente para comparecientes. No obstante, este valor ya sugiere una estructura discursiva bien definida para las víctimas. En el subcaso de la Costa Caribe, se observa una modularidad de 0.645 para comparecientes y 0.625 para víctimas, ambas superiores al valor agregado de 0.596, lo que indica una segmentación semántica más clara cuando se analizan los discursos por separado.

\begin{table}[!htb]
\centering
\begin{tabular}{lccc}
\hline
Subcaso & Todos & Comparecientes & Víctimas \\ \hline
Antioquia          & 0.616 & - & 0.616 \\ 
Casanare           & 0.489 & 0.408 & 0.533 \\ 
Costa Caribe       & 0.596 & 0.645 & 0.625 \\ 
Huila              & 0.337 & 0.302 & 0.426 \\ 
General            & 0.360 & 0.314 & 0.480 \\ \hline
\end{tabular}
\caption{Comparación de modularidad en redes de skipgramas según subcaso y rol de los participantes (víctimas y comparecientes).}
\label{tab:skipgramas_regiones_completo}
\end{table}

Este patrón sugiere que en estos dos territorios —particularmente en la Costa Caribe— existen diferencias significativas en la forma en que las víctimas y los comparecientes articulan sus relatos dentro del proceso judicial, reflejando narrativas más especializadas y cohesionadas al interior de cada grupo. Esta heterogeneidad discursiva ya se había anticipado en los análisis de sentimiento y de tópicos, y ahora se confirma desde una perspectiva estructural mediante el análisis de redes semánticas.

En contraste, para los subcasos de Casanare, Huila y el caso general, los valores de modularidad tienden a ser más bajos y las diferencias entre víctimas y comparecientes menos marcadas. Por ejemplo, en Casanare la modularidad pasa de 0.489 (general) a 0.533 (víctimas) y 0.408 (comparecientes), mientras que en Huila los valores son aún menores (0.337 general, 0.426 víctimas, 0.302 comparecientes), lo que sugiere una menor diferenciación discursiva entre los roles. De manera similar, el caso general muestra una modularidad de 0.360, que aumenta ligeramente a 0.480 para víctimas, pero sigue siendo más baja en comparación con los subcasos más estructurados.

Al comparar estos resultados con los obtenidos en el análisis de bigramas, se observa una tendencia similar: los subcasos con mayor modularidad (Antioquia y Costa Caribe) tienden a mantener esa coherencia tanto en bigramas como en skipgramas, aunque en estos últimos se capturan con mayor claridad las diferencias internas entre grupos. Esto refuerza la utilidad del análisis de skipgramas para revelar estructuras narrativas más profundas y complejas, especialmente en contextos donde la forma en que se expresan las memorias y responsabilidades varía de manera significativa entre los actores del conflicto.

\section{Discusión}

Este estudio presenta un abordaje computacional e interdisciplinar al Caso 03 de la Jurisdicción Especial para la Paz (JEP), enfocado en las ejecuciones extrajudiciales o \textit{falsos positivos}, mediante el análisis sistemático de los subtítulos de las audiencias públicas disponibles en video. La combinación de técnicas de procesamiento de lenguaje natural (PLN), análisis de redes semánticas y métodos de agrupamiento permitió explorar las narrativas y representaciones del conflicto desde múltiples dimensiones discursivas. Uno de los hallazgos centrales es la marcada intensidad del sentimiento negativo a lo largo de los discursos, aunque no necesariamente su mayor frecuencia en comparación con el sentimiento positivo. Esta distinción entre intensidad y recurrencia es clave para comprender la carga emocional presente en los testimonios, tanto de víctimas como de comparecientes.

El análisis regional reveló variaciones importantes en las dinámicas discursivas. En el subcaso de Antioquia no se observó una predominancia del sentimiento negativo, ni en términos de intensidad ni de frecuencia, lo que contrasta con la Costa Caribe, donde dicho sentimiento domina ambos aspectos cuando se analiza el corpus en su conjunto. Sin embargo, una segmentación más fina muestra que, en la Costa Caribe, esta tendencia negativa se concentra principalmente en las declaraciones de comparecientes, mientras que las víctimas exhiben un equilibrio mayor entre sentimientos positivos y negativos. Estas diferencias regionales también se reflejan en los resultados de modularidad: mientras en la mayoría de las regiones víctimas y comparecientes comparten partes relevantes del discurso, la Costa Caribe representa una excepción con una mayor polarización temática entre actores.

El análisis de tópicos emergentes a partir de redes de co-ocurrencia de bigramas y skipgramas permitió capturar la complejidad del conflicto armado colombiano. Los discursos se articulan alrededor de ejes como las estructuras y jerarquías militares, las vivencias dolorosas y traumáticas, pero también emergen elementos asociados al perdón, la restauración y la búsqueda de verdad. Este hallazgo sugiere que las audiencias públicas, además de ser espacios de memoria y justicia, funcionan como escenarios de resignificación narrativa, donde la coexistencia de dolor y agencia reparadora es posible.

A partir de este análisis, sustentado en análisis cuantitativo, es posible concluir que la realidad del conflicto armado colombiano está profundamente marcada por factores culturales, geográficos y contextuales. Estos elementos configuran la manera en que cada uno de los actores involucrados experimenta y comprende el conflicto. Este trabajo visibiliza algunas de esas diferencias y, al mismo tiempo, busca generar un efecto integrador que contribuya a construir una narrativa que refleje la diversidad, complejidad y los múltiples matices de esta historia.

Es fundamental destacar que el objetivo de estos análisis no es deshumanizar el discurso ni imponer una versión única o definitiva de los hechos. Por el contrario, se pretende aportar una herramienta de análisis que permita construir una visión más amplia e incluyente del conflicto, incorporando la mayor cantidad posible de variables, voces y perspectivas. Asimismo, se busca honrar a las víctimas que participaron en los procesos de justicia transicional, quienes, con valentía, compartieron relatos profundamente dolorosos. Estas historias, que nunca debieron ocurrir, deben ser documentadas con el mayor rigor y respeto para que las futuras generaciones conozcan su historia y trabajen activamente para que no se repita.

Desde el punto de vista metodológico, el estudio aporta una propuesta replicable y transparente que conjuga análisis cualitativo asistido por técnicas cuantitativas y visualización de datos. Se emplearon modelos de co-ocurrencia lingüística, medidas de modularidad para segmentación temática y análisis de sentimiento adaptado al español colombiano. Además, el proyecto pone a disposición la totalidad de sus scripts, grupos de palabras, gráficos y pruebas de hipótesis mediante un repositorio abierto en \url{https://github.com/aurreg/-Proyecto-JEP-}, lo que contribuye a la ciencia abierta y reproducible en el contexto de estudios de justicia transicional.

Sin embargo, el trabajo también enfrenta limitaciones importantes. El corpus analizado proviene únicamente de las audiencias públicas accesibles en línea, lo que puede sesgar los resultados y excluir testimonios relevantes no publicados o reservados. Asimismo, el análisis de sentimientos está restringido a interpretaciones léxicas de superficie y no contempla aspectos pragmáticos, contextuales o multimodales del discurso (como el tono, la gestualidad o las pausas). Por otra parte, el uso de palabras clave como proxy para identificar regiones geográficas puede no capturar la movilidad de actores o los desplazamientos forzados, relevantes en este tipo de conflicto.

Como líneas futuras de investigación, se propone enriquecer los modelos lingüísticos empleados mediante embeddings semánticos contextuales, como BERT en español, para mejorar la comprensión del discurso más allá de las co-ocurrencias simples. También sería pertinente incorporar variables sociodemográficas y judiciales de las y los comparecientes, para explorar correlaciones entre perfiles institucionales y estructuras discursivas. Finalmente, aplicar esta metodología a otros casos priorizados por la JEP o extenderla a contextos transicionales en otros países podría ofrecer una base comparativa valiosa para el estudio de memorias, responsabilidades y discursos sobre justicia y reparación. 

Por otro lado, esta metodología también se puede aplicar a otros cuerpos judiciales que tengan a cargo casos con múltiples y extensos medios de prueba (testimoniales, documentales, entre otros) para facilitar la comparación de discursos y la generación de una versión lo más cercana a la verdad, lo que ayudará a los jueces a poder determinar los hechos en diferentes casos. 

\section*{Statements and declarations}

The authors declare that they have no known competing financial interests or personal relationships that could have appeared to influence the work reported in this article.

During the preparation of this work the authors used ChatGPT-4-turbo in order to improve language and readability. After using this tool, the authors reviewed and edited the content as needed and take full responsibility for the content of the publication.

\bibliographystyle{apalike}
\bibliography{references.bib}

\end{document}